\pdfoutput=1

\documentclass[11pt]{article}

\usepackage[final]{acl}

\usepackage{times}
\usepackage{latexsym}

\usepackage{amsmath}

\usepackage{breqn}

\usepackage[T1]{fontenc}

\usepackage[utf8]{inputenc}

\usepackage{microtype}

\usepackage{inconsolata}

\usepackage{graphicx}
\usepackage{placeins}

\usepackage{tipa}
\usepackage{multirow}
\usepackage{cleveref}
\usepackage{dirtytalk}
\usepackage[disable]{todonotes}

\usepackage{newfloat}
\usepackage{caption}
\DeclareFloatingEnvironment[name=Box,within=section]{mybox}
\crefname{mybox}{Box}{Boxes}

\newcommand\blfootnote[1]{%
  \begingroup
  \renewcommand\thefootnote{}\footnote{#1}%
  \addtocounter{footnote}{-1}%
  \endgroup
}

\usepackage{booktabs}
\usepackage{colortbl}
\newcommand{\greyrule}{\arrayrulecolor{black!30}\midrule\arrayrulecolor{black}}
\usepackage{stackengine}

\newcommand\coderepo{https://github.com/hans/nnvb-public}
\title{Emergent morpho-phonological representations\\in self-supervised speech models}
\author{
Jon Gauthier\textsuperscript{1} \quad
Canaan Breiss\textsuperscript{2,3} \quad
Matthew Leonard\textsuperscript{1} \quad
Edward F. Chang\textsuperscript{1} \\
\textsuperscript{1}Department of Neurological Surgery, University of California, San Francisco \\
\textsuperscript{2}Department of Linguistics, University of Southern California \\
\textsuperscript{3}Center for Computational Language Sciences, University of Southern California\\
\texttt{jon@gauthiers.net} \quad
\texttt{cbreiss@usc.edu}
}

\usepackage{comment}

\begin{document}

\maketitle

\begin{abstract}
Self-supervised speech models can be trained to efficiently recognize spoken words in naturalistic, noisy environments. However, we do not understand the types of linguistic representations these models use to accomplish this task. To address this question, we study how S3M variants optimized for word recognition represent phonological and morphological phenomena in frequent English noun and verb inflections. We find that their representations exhibit a global linear geometry which can be used to link English nouns and verbs to their regular inflected forms.

This geometric structure does not directly track phonological or morphological units. Instead, it tracks the regular distributional relationships linking many word pairs in the English lexicon---often, but not always, due to morphological inflection. These findings point to candidate representational strategies that may support human spoken word recognition, challenging the presumed necessity of distinct linguistic representations of phonology and morphology.
\end{abstract}

\section{Introduction}

In everyday speech perception, humans transform a sequence of rapidly articulated words, often at a rate of several words per second, into coherent meanings.\blfootnote{Code to reproduce all results is available at \mbox{\url{\coderepo}}.}
Traditional psycholinguistic models attempt to explain this behavior by positing distinct representations of perceived speech at several linguistic levels: listeners processing individual phonetic segments in a word, along with the phonological, morphological, and semantic properties of a word. This transduction from low- to high-level representation happens through a series of recurrent computations %
\citep{mcclelland1986trace,gaskell1997integrating}.

While these models successfully explain a host of phenomena within the psycholinguistics of word recognition, they operate at a scale far below today's self-supervised speech models (S3Ms). Modern speech recognition algorithms that use S3Ms can make context-sensitive transcriptions of noisy acoustic input with an accuracy, efficiency, and scale not achievable with these psychologically motivated models. It is thus necessary to reconcile the theoretical commitments of earlier psycholinguistic models with the behavioral superiority of S3Ms.

Recent work has begun to study the representations of S3Ms trained on unlabeled speech data, asking to what extent they recover traditional psycholinguistic notions in phonetics, phonology, morphology, and syntax \citep{dunbar2022self,pasad2023comparative,pasad2023self,martin2023probing,sanabria2023analyzing,choi2024self,choi2025leveraging}. As a psycholinguistic account, this work departs from prior approaches by searching for model capacities which emerge as a solution to a training objective, rather than seeking to build in hypothesized structures.

However, it is not clear how these capacities of S3Ms relate---if at all---to the function of word recognition. Because S3Ms are trained on broad objectives unrelated to any specific linguistic task, their internal states plausibly serve multiple functions beyond recognizing words. As such, we do not know which components of their learned representations are \emph{necessary} for recognizing words, as opposed to predicting other aspects of the speech signal (from speaker identity and prosody to the identities of individual speech segments).

We present an analysis method and a series of controlled experiments on S3Ms, separating representations which are necessary for word recognition from those which emerge in service of more general objectives.
We first design a probing method to operationalize the notion of optimal word recognition. This probe targets a subspace of an S3M which serves to distinguish spoken words. In a series of experiments studying the activations within this word-optimal subspace, we discover a highly regular representation of speech sounds. We argue that this representation defies the cleanly separated levels of prior psycholinguistic models, cross-cutting morphological and simpler phonological distinctions, and instead tracks a higher-level pattern that unites multiple morphological inflections that all conform to a single phonological distribution rule. Our findings motivate a new candidate representational strategy that may serve spoken word recognition in human listeners.

\section{Background}

Like many languages, English exhibits regular sound patterns which reflect an interaction of phonological, morphological, and lexical constraints. For example, English words ending in [z], [s], or [\textipa{Iz}] can reflect plural nouns (NNS; e.g. \emph{dogs}, \emph{cats}) and third-person singular verbs (VBZ; e.g. \emph{runs}, \emph{barks}). Both of these inflections introduce a morpheme with the underlying phonological content /-z/ at the right edge of their base, which is realized as one of three surface forms (\textit{allomorphs}) depending on a simple set of distributional constraints, given in \Cref{box:phon-rules}.
\begin{mybox}
\setlength{\fboxsep}{1em}
\centering
\fbox{%
\parbox{0.85\linewidth}{
\begin{description}
    \item[/z/ $\rightarrow$ {[z]}] after voiced non-sibilant sounds (\emph{dogs} [\textipa{dOg\underline{z}}], \emph{runs} [\textipa{\*r\textturnv{}n\underline{z}}])
    \item[/z/ $\rightarrow$ {[s]}] after voiceless non-sibilant sounds (\emph{cats} [\textipa{kæt\underline{s}}], \emph{jumps} [\textipa{dZ\textturnv{}mp\underline{s}}])
    \item[/z/ $\rightarrow$ {[\textipa{Iz}]}] after sibilant sounds (\emph{dishes} [\textipa{dIS\underline{Iz}}], \emph{finishes} [\textipa{fInIS\underline{Iz}}])
\end{description}%
}
}
    \caption{Distributional constraints describing how the word-final /-z/ of English noun plurals and third-person verb inflections are realized in different surface forms [z], [s], and [\textipa{Iz}].}
    \label{box:phon-rules}
\end{mybox}

Of course, not all word-final instances of [z], [s], and [\textipa{Iz}] are generated by these morpho-phonological processes: consider monomorphemic words such as  \emph{haze}, \emph{fleece}, \emph{six}, and \emph{hearse}. Some of these words happen to end in sequences that are consistent with the constraints of \Cref{box:phon-rules}, while some do not: \emph{haze} and \emph{six} end in a sound matching the voicing of the preceding sound (consistent); \emph{fleece} and \emph{hearse} end in sounds which do not match the voicing of the preceding sound (inconsistent).

A combination of morphological, phonological, and lexical processes produce these word-final sounds. We exploit these multiple levels of patterning in a series of experiments on a model of word recognition, asking how representations at these levels are negotiated by the model in service of its objective.


\section{Methods}
We design experiments targeting four (non-exclusive) hypotheses about the linguistic representations in self-supervised speech models. From model activations computed on the word-final sounds [z], [s], and [\textipa{Iz}], our experiments evaluate:

\begin{description}
\item[Morphological sensitivity (\S\ref{sec:exp-morphology}):] A morphologically sensitive representation of [z] would contrast instances of the plural [z] (e.g. in \emph{daughters}) from instances of third-person singular [z] (in \emph{enters}); likewise for [s] and [\textipa{Iz}].
\item[Phonological sensitivity (\S\ref{sec:exp-allomorph}):] A phonologically sensitive representation of [z] would contrast instances where it surfaces as [z] (e.g. \emph{daughters}) from instances where it surfaces as [s] and [\textipa{Iz}] (e.g. \emph{lips} and \emph{cheeses}); and likewise for [s] and [\textipa{Iz}] in both noun plurals in verb inflections.
\item[Lexical sensitivity (\S\ref{sec:exp-false-friends}):] A lexically sensitive representation of [z] would contrast instances where the [z] is part of a lexical form (e.g. \emph{haze}) from instances where it is inflectional (e.g. \emph{daughters}); and likewise for [s] and [\textipa{Iz}].
\item[Distributional sensitivity (\S\ref{sec:exp-forced-choice}):] A representation sensitive to the distributional constraints given in \Cref{box:phon-rules} would contrast sounds [z], [s] and [\textipa{Iz}] in contexts where they are consistent with these rules (e.g. \emph{haze}) from contexts where they are inconsistent (e.g. \emph{fleece}).
\end{description}

We address these questions using two types of models: a self-supervised speech model trained with a general contrastive learning objective, and a fine-tuned variant trained specifically for word recognition. By comparing answers to the above questions across these two models, we can assess which levels of linguistic representation are actually necessary for performing word recognition.


\subsection{Base model}
\label{sec:base-model}

We begin our analyses with the Wav2Vec2 Base model \citep[herein Wav2Vec]{baevski2020wav2vec}\footnote{\href{https://huggingface.co/facebook/wav2vec2-base}{\texttt{huggingface.co/facebook/wav2vec2-base}}}, a self-supervised Transformer model of raw audio, which was trained on 960 hours of unlabeled data from the LibriSpeech corpus \citep[publicly available under a CC-BY 4.0 license]{panayotov2015librispeech}. This Transformer model takes an audio waveform as input and produces frame representations $x_\ell^{(t)}$, a sequence of 768-dimensional vectors each spanning 20 ms of audio beginning at time $t$, arranged in a series of model layers $\ell$. While this variant of Wav2Vec was trained with entirely unlabeled audio, without phone- or word-level annotations, these models still accumulate detailed high-level representations of the linguistic input in individual frames \citep{pasad2023self}.

\subsection{Word probe model}
\label{sec:probe-model}

What is the subspace of these Wav2Vec representations which optimally contrasts spoken words? We target this subspace by defining a linear projection on Wav2Vec's representation $x_\ell^{(t)}$ at frame $t$ and layer $\ell$ onto a vector $z_\ell^{(t)}$:
\begin{equation}
    z_\ell^{(t)} = W_z x_\ell^{(t)} \label{eqn:wav2vec-frame}
\end{equation}
This probe is optimized with a contrastive learning objective. For each frame $t$ within the span of a word $j$, we take all other frames $t^+$ spanned by other tokens of $j$ as positive examples, and all other frames $t^-$ spanned by tokens of distinct words as negative examples. We minimize a hinge loss, describing the separation in cosine distance between a frame and its positive and negative examples with  margin parameter $m$:
\begin{dmath}
    \mathcal L(t) = \max\left(0, m + \cos(z_\ell^{(t)}, z_\ell^{(t^+)}) - \cos(z_\ell^{(t)}, z_\ell^{(t^-)})\right)
    \label{eqn:loss}
\end{dmath}
We train this probe on 100 hours of word-aligned audio from LibriSpeech from the split \texttt{train-clean-100}.\footnote{Alignments available in \citet{lugosch2019librispeech}.} This model is trained to convergence separately for each layer $\ell$ of the Wav2Vec2 base model (12 layers in total). Further optimization details are included in \Cref{appendix:training}.

We take the activations of this model $z_\ell^{(t)}$ to be an optimal subspace of Wav2Vec for performing word-level contrast.
By analyzing these activations, we can understand which aspects of the speech input are exploited for word recognition. By comparing these activations to the original Wav2Vec activations $x_\ell^{(t)}$, we can identify what information is present in the input but discarded by the model when performing word-level contrast.

\subsection{Acoustic word embeddings}
\label{sec:word-embeddings}

For all word tokens $w_j$ in the word-annotated LibriSpeech corpus, we compute a fixed-length word representation by averaging across the $N_j$ frame representations between the word's onset time $o_j$ and offset \citep{sanabria2023analyzing,pasad2023self}.\footnote{While our experiments center around these word-level embeddings, this method is not critical to our results. \Cref{sec:phoneme-pooling} shows that our results hold under a different embedding and analogy method, using representations pooled within individual phonemes rather than across entire words.}
\begin{equation}
    f_\ell(w_j) = \frac{1}{N_j} \sum_{t=0}^{N_j - 1} z_\ell^{(o_j + t)}
    \label{eqn:wordrep}
\end{equation}
These frame representations may correspond to the hidden states of Wav2Vec at layer $\ell$, $x_\ell^{(t)}$, or the word probe $z_\ell^{(t)}$.

\subsection{Experiments}
\label{sec:experiments}
\newcommand\librispeechNumTokens{\ensuremath{334,008}}

\begin{table}[t]
  \centering
  \resizebox{\linewidth}{!}{
  \begin{tabular}{ p{1.7cm}lll }
    \toprule
    Inflection & Allomorph & Base & Inflected \\
    \midrule
    \multirow{3}{*}{\shortstack[c]{Noun plural\\(NNS /z/)}} & [z] & \emph{daughter} & \emph{daughters} \\
    & [s] & \emph{lip} & \emph{lips} \\
    & [\textipa{Iz}] & \emph{cheese} & \emph{cheeses} \\
    \midrule
    \multirow{3}{*}{\shortstack[c]{Verb 3SG\\(VBZ /z/)}} & [z] & \emph{give} & \emph{gives} \\
    & [s] & \emph{exist} & \emph{exists} \\
    & [\textipa{Iz}] & \emph{please} & \emph{pleases} \\
    \bottomrule
  \end{tabular}
  }
  \caption{Examples of unambiguous regular inflections.}
  \label{tbl:inflections}
\end{table}
Our experiments use the vector analogy method of \citet{mikolov2013distributed} to study the model's ability to generalize its representations of the relevant speech sounds [z], [s], and [\textipa{Iz}] to lexically, phonologically, and morphologically contrasting strings.

Our analogy trials link two pairs of words, each of which are phonologically identical but for a final [z], [s], or [\textipa{Iz}]. \Cref{tbl:inflections} gives examples of such pairs consistent with either noun pluralization or verb inflection. An example analogy is as follows:

\begin{center}
\emph{shirt} [\textipa{S\textrhookschwa t}]\ : \emph{shirts} [\textipa{S\textrhookschwa t}s]\ :: \emph{cheese} [\textipa{\textteshlig{}iz}]: \underline{\hspace{1cm}}
\end{center}

We implement this analogy with vector algebra. For random samples of word tokens $a = \textit{shirt}, b = \textit{shirts}, c = \textit{cheese}$, we compute word token representations $f_\ell(a_i), f_\ell(b_i), f_\ell(c_i)$, following \Cref{eqn:wordrep}. We randomly draw token word embeddings $a_i, b_i, c_i$ and calculate:
\begin{equation}
\hat d_{i} = b_{i} - a_{i} + c_{i}
\label{eqn:analogy}
\end{equation}
For each trial, we compute the cosine distance between the predicted vector $\hat d_i$ and all word embeddings $f_\ell(w_j)$ computed across the LibriSpeech corpus. We average these predicted distances $d_i$ across instances of the given analogy, and use these to compute a single nearest-neighbors ordering over all word tokens. This yields a single ranking over all word tokens for each analogy structure $(a, b, c)$.

We evaluate with a rank metric: in the nearest-neighbors list for a given analogy, we find the position of the target word $d$. This rank value ranges from 0 to \librispeechNumTokens{} (the number of word tokens in the dataset).
We define a random baseline performance for an analogy pair $(a, b, c, d)$ by sampling a random counterfactual pair of words $(\overline a, \overline b)$ and using the same method to compute the rank of $d$ in the neighbors of predicted vectors $\overline b_i - \overline a_i + c_i$.

\subsection{Experimental design}
\newcommand\nunambignouns{753}
\newcommand\nunambigverbs{52}
We select \nunambignouns{} frequent English nouns and \nunambigverbs{} verbs which were present in the LibriSpeech dataset and are unambiguous with respect to part of speech: that is, their inflected form is clearly a noun or a verb.\footnote{This is not true of most English nouns: for example, the form \emph{stacks} can be either the plural of the noun \emph{stack} or the third-person singular of the verb \emph{to stack}. We also exclude words which are technically unambiguous, but which have base or inflected forms that are homophonous with a monomorpheme: for example,  \emph{patients}--\emph{patience}; \emph{knows}--\emph{nose}. Additional details on word selection are in \Cref{appendix:words}.} While these unambiguous pairs share no actual morphological relationship (they instantiate an unambiguous plural noun or verb inflection), they do share a phonological relationship, all exhibiting inflected forms with an underlying word-final -/z/.

\section{Results}

\newcommand\randomChanceRank{\ensuremath{94,306}}
\newcommand\forcedChoiceN{\ensuremath{35}}

\newcommand\ttestNounVerb{\ensuremath{t=-4.01, p<10^{-4}}}
\newcommand\ttestWavVecWord{\ensuremath{t=4.43,p<10^{-5}}}

\renewcommand\stacktype{L}
\begin{table*}[t]
    \resizebox{\linewidth}{!}{
    \begin{tabular}{lp{6cm}p{8.2cm}p{3cm}}
        \toprule
         & Method & Example & Answer \\
        \midrule
        \S\labelcref{sec:exp-morphology} & Test analogy across \textbf{morphological} categories \mbox{(NNS$\rightarrow$VBZ, VBZ$\rightarrow$NNS)} & \raisebox{-2.6ex}{\stackon{/z/ --- NNS}{\emph{daughter} : \emph{daughters}} \raisebox{2.6ex}{::} \stackon{/z/ --- VBZ}{\emph{own} : \emph{owns}}} & Invariant \\
        \greyrule
        \S\labelcref{sec:exp-allomorph} & Test analogy across \textbf{allomorphs} \mbox{([z], [s], [\textipa{Iz}])} & \raisebox{-3ex}{\stackon{[z]}{\emph{daughter} : \emph{daughters}} \raisebox{3ex}{::} \stackon{[s]}{\emph{lip} : \emph{lips}}} & Invariant \\
        \greyrule
        \S\labelcref{sec:exp-false-friends} & Test analogy to/from false inflections & \emph{daughter} : \emph{daughters} :: \emph{beside} : \emph{besides} & Invariant \\
        \greyrule
        \S\labelcref{sec:exp-forced-choice} & Forced-choice analogy evaluation for \textbf{phonological consistency} & \raisebox{-1.9ex}{\emph{daughter} : \emph{daughters} :: \emph{bay} : $\left\{\begin{array}{l}\textit{bays}\text{ (consistent)}\\\textit{base}\text{ (inconsistent)}\end{array}\right\}$} & Prefers sounds consistent with \Cref{box:phon-rules} \\
        \bottomrule
    \end{tabular}
    }
    \caption{Summary of experimental designs and results on the word probe.}
    \label{tbl:summary}
\end{table*}

\begin{figure}[t]
  \centering
  \includegraphics[width=\linewidth]{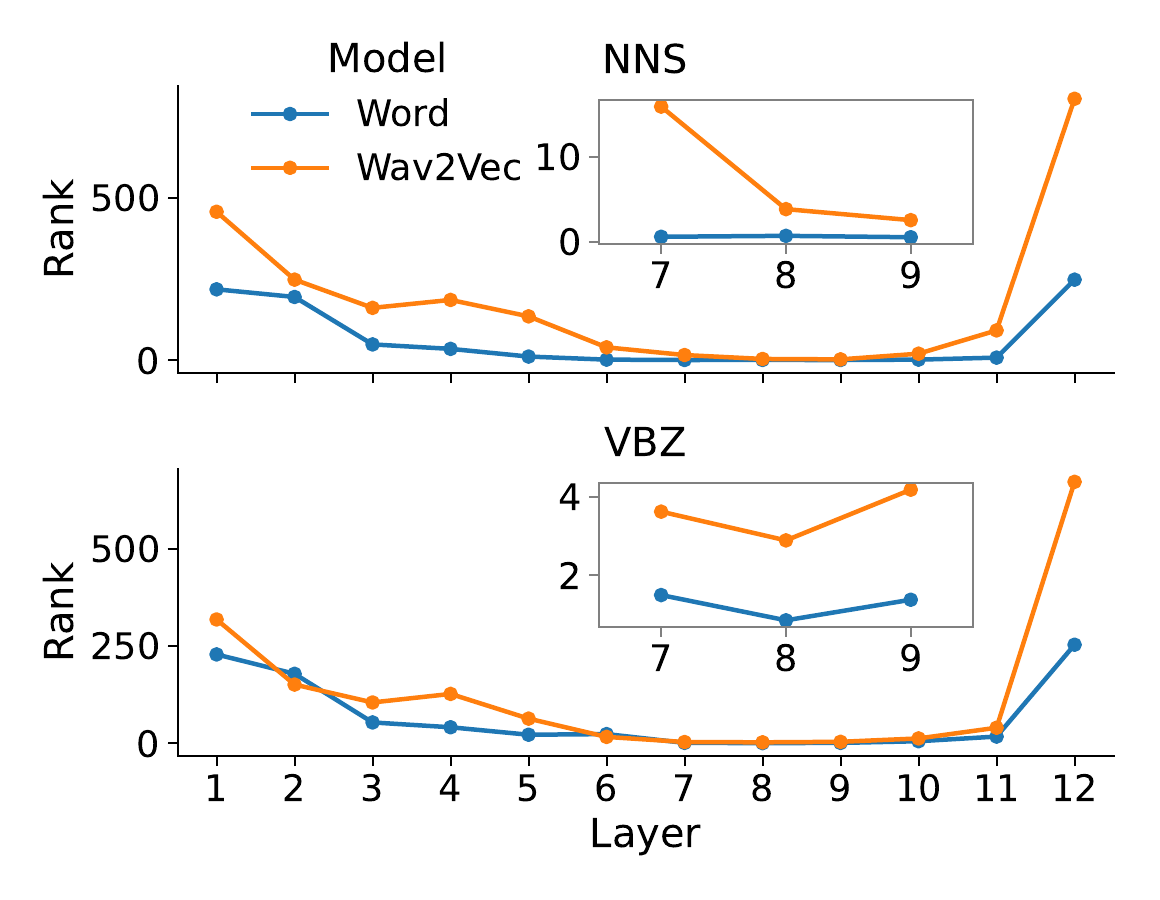}
  \caption{Per-layer analogy rank performance (lower is better) for regular noun and verb inflections. Random chance is \randomChanceRank. Insets show zoom on layers 7--9.}
  \label{fig:layer-wise}
\end{figure}
We first evaluate the overall capacity of each layer of our word probe, along with Wav2Vec, to solve analogy tasks across these English noun and verb inflections.
\Cref{fig:layer-wise} plots results at each layer; model predictions exceed random chance (rank $=~ \randomChanceRank$) for all inflections at all layers.

These layer-wise results mirror the typical pattern of abstract feature encoding in S3Ms, with maximal performance in intermediate layers and a substantial decrease in performance in final layers closest to the model's prediction head. The performance of Wav2Vec peaks in the 8th and 9th layer, matching the peak location of phonetic encoding found in prior studies \citep{pasad2023comparative}.

\begin{figure}[t]
    \centering
    \includegraphics[width=0.85\linewidth]{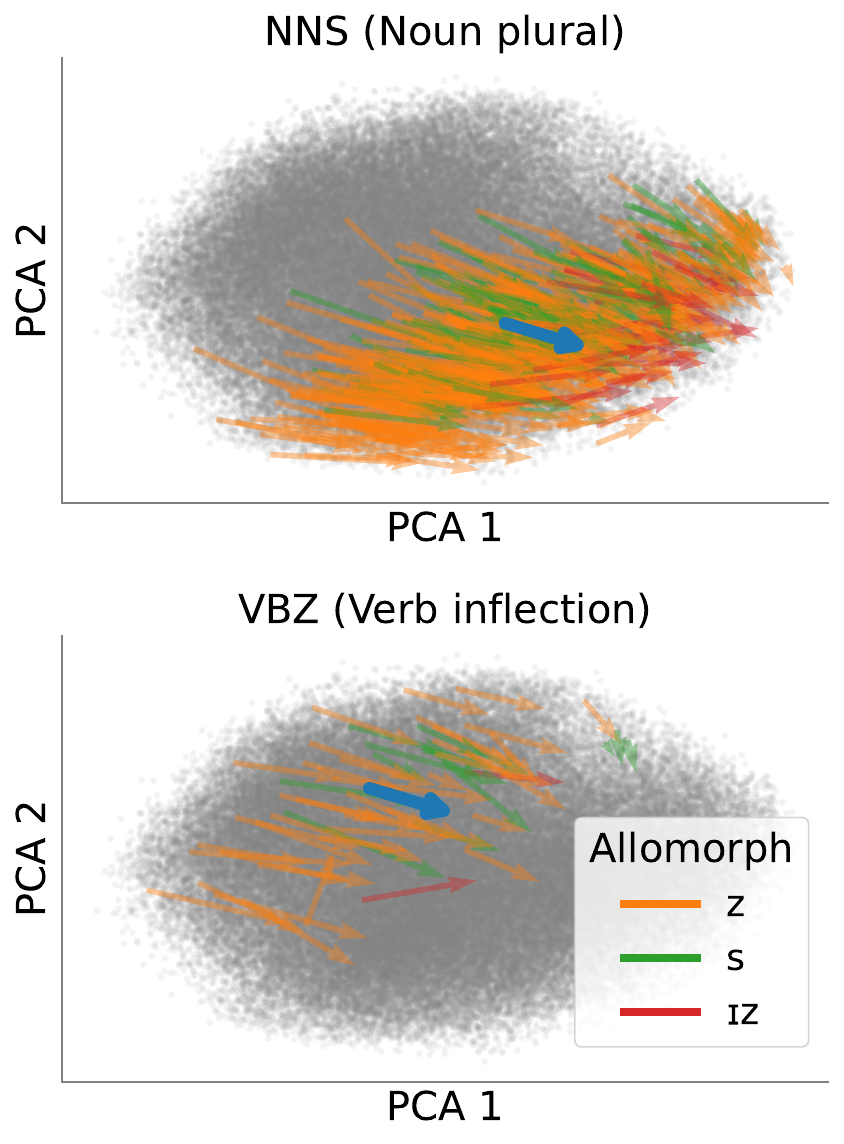}
    \caption{Difference vectors from base forms $a$ to inflected forms $b$, computed on layer 8 of the word probe and projected into the first two principal components of embedding space for all words in our study. Bold blue line shows mean direction vector; gray dots show a random sample of word embeddings.}
    \label{fig:pca}
\end{figure}

Both Wav2Vec and the word probe thus exhibit a global linear geometry that links inflected nouns and verbs to their base forms. This geometry is visible in the principal component space of all acoustic word embeddings, visualized in \Cref{fig:pca}.

What kind of linguistic information is captured by these difference vectors? By design, the analogy trials we construct frequently link source and target pairs which are mismatched in morphological category (noun and verb inflections) and phonological detail (the particular allomorphs involved). The following experiments (summarized in \Cref{tbl:summary}) ask whether the particular morphological and phonological relationship between the source and target pairs affect the success of this analogy task. If analogy succeeds despite morphological and phonological mismatches, this indicates that the difference vectors are not dependent on distinctions at these levels.
We present experiments on the highest-performing layer in Wav2Vec and the word probe (layer 8).
We focus on high-level mean rank results in the main text, with more detailed quantitative analyses provided in \Cref{sec:regression} and \Cref{sec:extended-results}.

\subsection{Evaluating morphological sensitivity}
\label{sec:exp-morphology}
\begin{figure}[t]
    \centering
    \includegraphics[width=\linewidth]{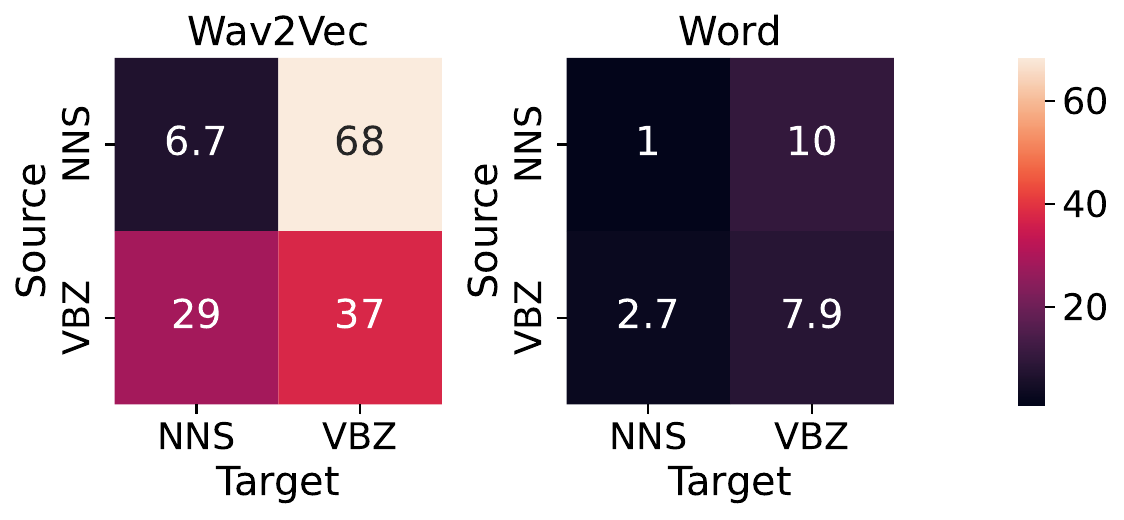}
    \caption{Mean rank values (lower is better) from analogy within and between noun/verb inflections on the Wav2Vec baseline and word probe.}
    \label{fig:transfer-morph}
\end{figure}

We first evaluate the model's ability to generalize within and across morphological categories, from plural noun inflections to both 1) other plural noun inflections and 2) third-person verb inflections, and vice-versa. If the representation supporting overall analogy performance is primarily morphological, this latter evaluation should fail, since by design there is no shared morphological relationship between our unambiguous noun pairs and verb pairs. If the representation does not depend on morphological facts, we should see success in both cases.

\Cref{fig:transfer-morph}\todo{it would be great to have example words in all of these, if possible / space} shows the results of this analogy evaluation. In each heatmap, the diagonal elements show the average rank values resulting from analogies within-inflection (e.g. \emph{shirt}/\emph{shirts} to \emph{cheese}/\emph{cheeses}), while the off-diagonal elements show the results of performing analogies between inflections (e.g. \emph{shirt}/\emph{shirts} to \emph{enter}/\emph{enters}).

We first examine the results of the word probe (right heatmap). By comparing the NNS$\rightarrow$NNS cell (top left) and VBZ$\rightarrow$VBZ cell (bottom right), we can see a main effect of morphology: while both categories perform far above chance, nouns are better predicted than verbs (\ttestNounVerb).\footnote{Error analysis suggests that this main effect stems from differences in the size of noun and verb morphological paradigms. English nouns have a very small morphological paradigm primarily involving pluralization, while verbs have a larger paradigm (3SG, past, participle, gerund). A common error in 3SG verb prediction is selecting the wrong in\-flec\-tion within this paradigm, such as predicting \emph{pleasing} instead of \emph{pleases} from \emph{please}.}
The off-diagonal cells test analogy from noun plural inflections to verb inflections and vice-versa. This evaluation also performs well above chance, though we see a similar superiority for predicting noun inflections over verb inflections.

These results show a highly restricted role of morphology within the word probe's computations. This contrasts with the results of the same evaluation applied to Wav2Vec's internal states, shown in the left panel of \Cref{fig:transfer-morph}.
Wav2Vec performs substantially worse overall (\ttestWavVecWord), and also shows a much larger sensitivity to morphological contrasts: while the word probe shows an average rank difference of mismatching morphology of 7.1, the same metric in the Wav2Vec results is 34.7.
The word probe thus supports computations which are relatively invariant to morphological contrasts.

\subsection{Evaluating phonological sensitivity}
\label{sec:exp-allomorph}
\begin{figure*}[t]
    \centering
    \includegraphics[width=0.9\linewidth]{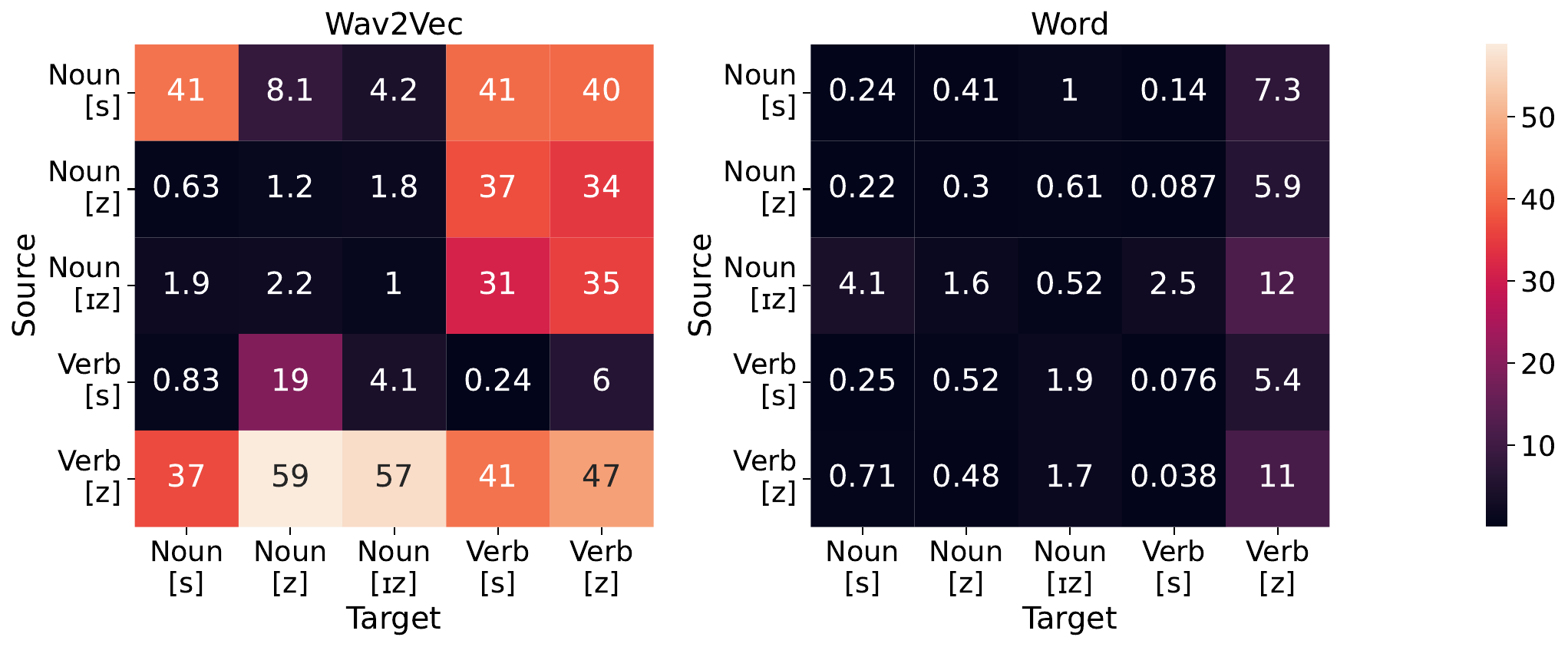}
    \caption{Analogy within and between allomorphs of noun and verb inflections. Heatmaps show an average rank metric (0 is best; random guessing is \randomChanceRank). The word probe in the right panel exhibits strongly reduced sensitivity (i.e., improved performance) to allomorphic contrasts.}
    \label{fig:transfer-allomorph}
\end{figure*}

We next addressed our question about sensitivity to phonological contrasts within and between each inflectional category. We did this by splitting the previous unambiguous noun--verb experiment based on the particular allomorph ([z], [s], or [\textipa{Iz}]) involved in both the base and target pair.

\Cref{fig:transfer-allomorph} shows the results of this evaluation applied to our probe model. Each cell value indicates the accuracy of the model in generalizing from some particular allomorph base to a different allomorph target pair. The Wav2Vec model (left panel) shows strong sensitivity to phonological and morphological distinctions: while analogy performance is almost as good as that of the word probe model for particular cases (e.g. VBZ [s] $\rightarrow$ NNS [s], \emph{exist} : \emph{exists} :: \emph{lip} : \emph{lips}), the majority of mappings show degraded performance (e.g. VBZ [s] $\rightarrow$ NNS [\textipa{z}], \emph{exist} : \emph{exists} :: \emph{daughter} : \emph{daughters}).

In contrast, the word probe model (right panel) is relatively stable across both phonological and morphological differences between the source and target of an analogy. We still see moderate effects of morphological identity (in particular, the row-wise effect of drawing NNS [\textipa{Iz}] as a source, and the column-wise effect of generalizing to VBZ [z] as a target), but the scale of this degradation is far smaller than that of the Wav2Vec evaluation.

\subsection{Evaluating lexical sensitivity}
\label{sec:exp-false-friends}

Our results show that Wav2Vec's encoding of word-final [z], [s], and [\textipa{Iz}] is sensitive to both the morphological and phonological conditions distributing these sounds. These details are discarded, however, by a model optimized for word recognition.
We next ask if this pattern extends beyond the morphological phenomena of noun plurals and verb inflections, to distinct \emph{lexical} sources of word-final [z], [s], and [\textipa{Iz}].
\begin{table}[t]
    \centering
    \begin{tabular}{ll|ll}
        \toprule
        Base & Inflected & Base & Inflected \\
        \midrule
        backward & backwards & beside & besides \\
        the ([ði]) & these & though & those \\
        \bottomrule
    \end{tabular}
    \caption{False friends of the most frequent allomorph, [z].}
    \label{tbl:false-friends}
\end{table}
We derive ``false friend'' word pairs, which on their surface appear just like English noun and verb inflections: the ``inflected'' form is the concatenation of a ``base'' form with one of the sounds [z], [s], or [\textipa{Iz}], obeying the rules given in \Cref{box:phon-rules}. \Cref{tbl:false-friends} shows examples of such false friends attested in English.\footnote{We exclude all true nouns and verbs as possible false friends, since any analogy evaluation relating real nouns and verbs would confound the phonological ``false friend'' relationship with other morphological relationships.}

\begin{figure}[t]
    \centering
    \includegraphics[width=0.9\linewidth]{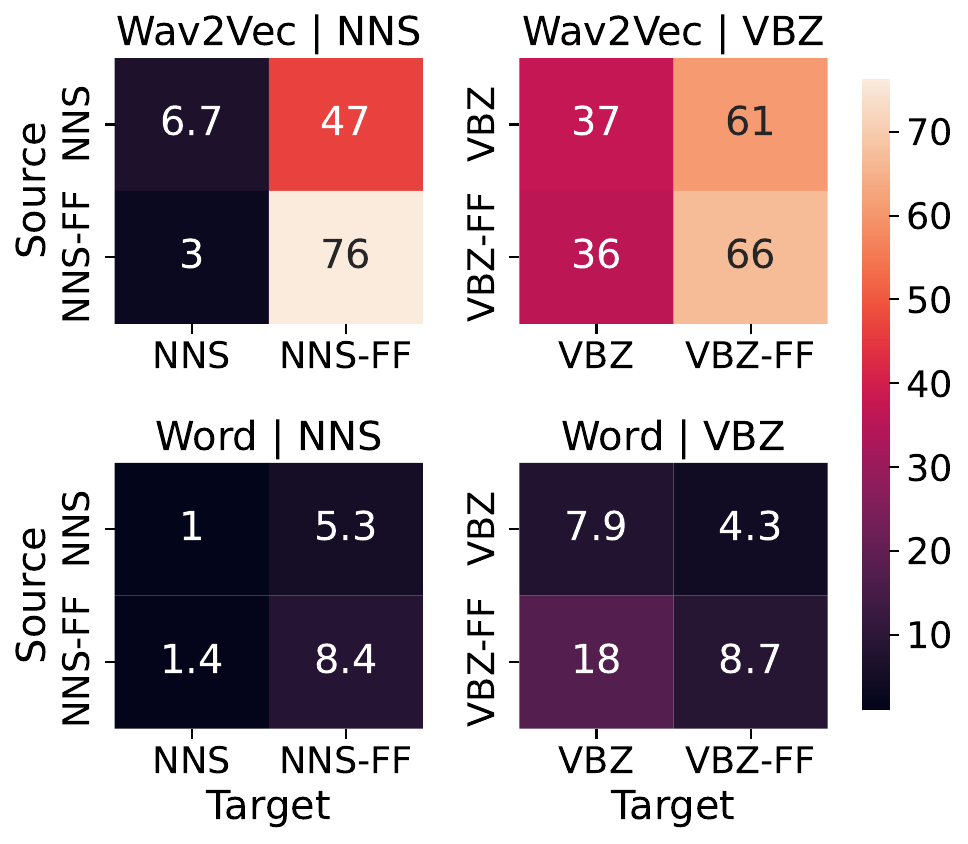}
    \caption{Analogy to false-friend (FF) inflection pairs. Word probe (bottom panels) shows reduced sensitivity to whether or not sounds are participating in true morphological inflections.}
    \label{fig:false-friends}
\end{figure}

\Cref{fig:false-friends} evaluates how difference vectors computed from true noun and verb inflections perform in predicting in false friend (FF) items, and vice versa, in both Wav2Vec and the word probe. The top-left cell of each panel indicates the analogy performance within valid noun and verb inflections; note that these values correspond to the diagonal values of \Cref{fig:transfer-morph}. The off-diagonals indicate the ability of the model to compute analogies linking valid inflections with false friends (for example, the top right cell of left panels, NNS$\rightarrow$NNS-FF, tests analogies such as \emph{shirt} : \emph{shirts} :: \emph{though} : \emph{those}).

The Wav2Vec model (top panels) shows a strong sensitivity to real versus false-friend inflections, as is visible on the diagonals of the heatmaps: noun analogy prediction degrades from a rank of 6.7 to 76, and verbs from 37 to 66. In contrast, we see a much smaller change in the word probe (lower panels), with a change in rank outcome in nouns from 1 to 8.4 and in verbs from 7.9 to 8.7.

This suggests that the word probe is relatively insensitive to the contrast between lexical and morphological sources of word-final [z], [s], and [\textipa{Iz}].
It must instead rely on the phonological relationship between these words' base and inflected forms.

\subsection{Evaluating distributional sensitivity}
\label{sec:exp-forced-choice}

But what kind of phonological relationship  is encoded in these difference vectors linking spoken words?
The allomorphy results of \Cref{fig:transfer-allomorph} show that the word probe model is apparently invariant to distinctions in English speech sounds which are contrastive: sounds which \emph{must} be distinctly represented in order to distinguish English words.

Concretely, consider the minimal pair of \emph{bays} and \emph{base}, both of which can be derived by the addition of a single sound to the word \emph{bay}.\footnote{The attentive reader will also note that these words may contrast in vowel length. This confound is addressed in \Cref{sec:fc-length-confound}.} While \emph{bays} is consistent with the rules given in \Cref{box:phon-rules}---it contains a word-final [z] following a voiced sound---\emph{base} is not consistent, placing a word-final [s] after a voiced sound. We have seen that the word probe is largely insensitive to the particular word-final sounds in these words. However, it is still possible that the word probe is sensitive to the \emph{distributional constraints} of these sounds: whether or not they are consistent with the kind of rules shown in \Cref{box:phon-rules}.

%
In a final experiment, we select \forcedChoiceN{} word triples (\Cref{tbl:fc-materials}) exhibiting the following property: the ``base'' is a substring of both other words; the consistent word is the concatenation of the base with a [s], [z], or [\textipa{Iz}], following the distributional constraints of \Cref{box:phon-rules}; the inconsistent word is the concatenation of the base with a sound not following these constraints.\footnote{Some of the phonologically consistent items are also morphologically related to the base (\emph{bay}--\emph{bays} is a plural inflection), while others have no relationship at all (\emph{knew}--\emph{news}).}
We perform a forced-choice analogy evaluation, mapping difference vectors from real noun and verb inflections onto the base word, e.g.:
\begin{center}
    \emph{lip} : \emph{lips} :: \emph{bay} : $\left\{ \begin{array}{l}\textit{bays}\text{ (consistent)}\\\textit{base}\text{ (inconsistent)}\end{array} \right\}$
\end{center}
\begin{figure}[t]
    \centering
    \includegraphics[width=0.75\linewidth]{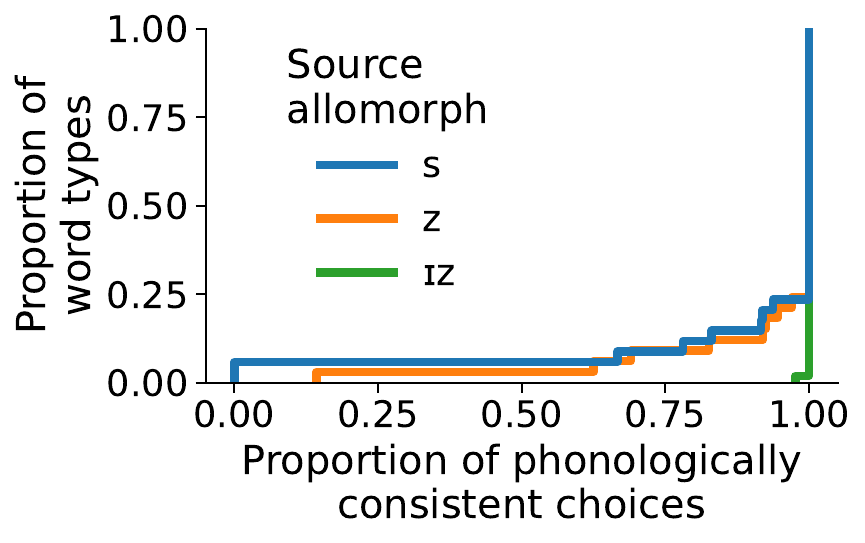}
    \caption{Cumulative distribution of preferences for the item obeying distributional constraints (\Cref{box:phon-rules}) in forced-choice evaluations on 35 target pairs. The majority of forced-choice evaluations predict the consistent item 100\% of the time.}
    \label{fig:forced-choice}
\end{figure}
We measure the probability that the model predicts a vector $\hat d_i$ (\cref{eqn:analogy}) whose nearest neighbor is a token of the phonologically consistent form \emph{bays} rather than the inconsistent form \emph{base}.
We find that the word probe's difference vectors overwhelmingly point to the consistent option of each forced-choice item, with a majority of items showing an absolute preference for consistency (\Cref{fig:forced-choice,fig:fc-barplot}).
This suggests that the word probe representations are not completely invariant to contrasts between these sounds. Instead, they capture the abstract distributional constraints that give rise to [z], [s], and [\textipa{Iz}] in regular noun and verb inflections.

\subsection{Regression evaluation}
\label{sec:regression}
\begin{figure}[t]
    \centering
    \includegraphics[width=0.65\linewidth]{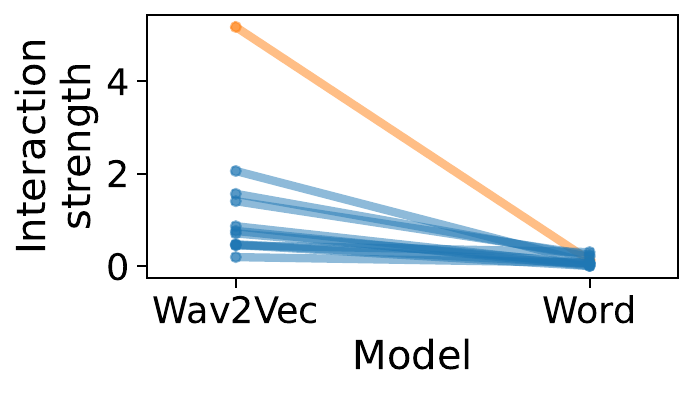}
    \caption{Interaction strength (a regression-based measure of sensitivity to morphological and allomorphic distinctions) in Wav2Vec and the word probe.}
    \label{fig:interaction-strength}
\end{figure}
Previous sections claimed that the word probe showed a substantial reduction in its sensitivity to morphological (\cref{sec:exp-morphology,sec:exp-false-friends}) and phonological (\cref{sec:exp-allomorph}) contrasts, relative to the Wav2Vec baseline model. We now quantitatively test these claims with a linear regression, predicting the rank outcome of individual analogy trials as a function of the allomorphs and morphological categories involved, along with nuisance predictors:
\begin{align}
    \texttt{rank} \sim\,& \texttt{allomorph\textunderscore{}from} \times \texttt{inflection\textunderscore{}from } \times \notag\\
        & \texttt{allomorph\textunderscore{}to} \times \texttt{inflection\textunderscore{}to } + \notag\\
        & \texttt{from\textunderscore{}frequency} + \texttt{to\textunderscore{}frequency}
    \label{eqn:regression}
\end{align}
where \texttt{inflection} features are categorically coded in $\{\texttt{NNS}, \texttt{VBZ}\}$, \texttt{allomorph} features are categorically coded in $\{\texttt{s}, \texttt{z}, \texttt{Iz}\}$, and \texttt{frequency} features are log-transformed frequencies from Worldlex \citep{gimenes2016worldlex}. This model is fit independently on the results of Wav2Vec and the word probe, at the same model layer studied in previous sections.

The interaction coefficients  of these regressions capture how sensitive the models are to matches or mismatches between the morphology and allomorphy of source and target pairs. \Cref{fig:interaction-strength} tracks the absolute value of each interaction coefficient explaining the performance of the Wav2Vec model and word probe.
For example, the strongest interaction coefficient in the Wav2Vec fit (left point of the orange line in \Cref{fig:interaction-strength}) captures the large difference between noun-to-verb and verb-to-verb analogies in the right column of \Cref{fig:transfer-morph}. This coefficient is greatly reduced in the word model (right end of orange line), along with that of all other interaction effects.

This reduction in interaction effects is consistent with our earlier findings: optimizing a model for word recognition produces representations with substantially reduced sensitivity to both allomorphic and morphological variation.
\Cref{appendix:regression} presents the full results of the regression evaluation.

\section{Discussion}

This paper used speech representations computed from self-supervised speech models (S3M) to ask: what kinds of representations are necessary for recognizing words?
We studied how an S3M variant optimized for word recognition, the word probe, negotiates phonological, morphological, and lexical levels of representation to serve this objective.

Our experiments show that the model represents an abstract phonological regularity in English: a highly frequent \emph{distributional regularity} (\Cref{box:phon-rules}) governs word-final [s], [z], and [\textipa{Iz}]. This knowledge is not strongly conditioned by morphology (\cref{sec:exp-morphology,sec:exp-false-friends}) or phonological relationships between allomorphs (\cref{sec:exp-allomorph}), and instead tracks a morpho-phonological process governing the sounds which form English noun plurals and present tense verb inflections (\cref{sec:exp-forced-choice}).
This generalization emerges early in the model (as early as layer 1) and peaks in intermediate layers (\cref{fig:layer-wise}). In the word probe, this knowledge is encoded as a global linear translation, prominent in the principal components of the embedding space (\cref{fig:pca}).

Our results add to prior probing studies of self-supervised speech models. \citet{pasad2023comparative} found that intermediate layers 8 and 9 of Wav2Vec most prominently encoded phonetic information. Our word probe results show that these layers also contain a linear subspace capable of more abstract phonological reasoning; on the surface, however, these representations are indeed sensitive to phonetic contrasts. \citet{choi2025leveraging} demonstrate that Wav2Vec's internal states retain sub-phonemic and allomorphic information; we confirm this allomorphic sensitivity in \Cref{sec:exp-allomorph}, and show that optimizing for an objective of word recognition effectively erases much of that allophonic information (\Cref{fig:transfer-morph}).

Our results also complement existing psycholinguistic models of spoken word recognition \citep[inter alia]{mcclelland1986trace,norris1994shortlist,gaskell1997integrating}, which largely were designed to explain small-scale phenomena of single-word recognition. Due to this data scale, these past models required strong inductive biases about the kinds of intermediate linguistic computations necessary for word recognition. In contrast, by combining large amounts of naturalistic data, low-bias neural network models, and controlled experiments, we can uncover novel hypotheses about the computations underlying spoken word recognition.

This kind of level-crossing speech representation is broadly consistent with some alternative distributed models of spoken word recognition in psycholinguistics, which emphasize how apparent rule-based morphological relationships can be modulated by linguistic knowledge at other levels of representation \citep{gonnerman2007graded}. It is also compatible with a view in the theory of Distributed Morphology, known as the \emph{separation hypothesis}, which similarly argues that representations of groups of phonologically distributed sounds (such as the set [z], [s], and [\textipa{Iz}]) may be disconnected from the their original morphological sources \citep{embick2022morphology}.

\section*{Acknowledgments}

We thank Connor Mayer, David Embick, members of the Chang Lab at UCSF and the Stanford GLySN lab, and the audience at the Bay Area Language Processing Interest Group for feedback on this work. Generative AI tools (ChatGPT 4o) were used to improve the wording and clarity of the manuscript.

\section*{Limitations}

This paper focuses on a small (albeit highly frequent) set of morphological and phonological phenomena in English. Future work should ask whether this linear geometry applies to other English phenomena with dual patterning of morphology and phonology --- for example, word-final \emph{-er}, which is generated by both comparative (\emph{bigger}) and agentive (\emph{buyer}) inflections. We might also use multilingual or monolingual non-English models to verify that the patterns revealed in our experiments are truly a generalization resulting from exposure to English phonology specifically.

Our analyses are limited to a very strict assumed geometric relationship between word forms (linear translation). Our findings may be sensitive to this geometric assumption, yielding both false positives (on the sensitivity of Wav2Vec to morphology and phonology) and false negatives (on the lack of sensitivity of the word probe model).

\bibliography{mybib}

\begin{thebibliography}{20}
\providecommand{\natexlab}[1]{#1}

\bibitem[{Baevski et~al.(2020)Baevski, Zhou, Mohamed, and Auli}]{baevski2020wav2vec}
Alexei Baevski, Yuhao Zhou, Abdelrahman Mohamed, and Michael Auli. 2020.
\newblock wav2vec 2.0: A framework for self-supervised learning of speech representations.
\newblock \emph{Advances in neural information processing systems}, 33:12449--12460.

\bibitem[{Choi et~al.(2024)Choi, Pasad, Nakamura, Fukayama, Livescu, and Watanabe}]{choi2024self}
Kwanghee Choi, Ankita Pasad, Tomohiko Nakamura, Satoru Fukayama, Karen Livescu, and Shinji Watanabe. 2024.
\newblock Self-supervised speech representations are more phonetic than semantic.
\newblock In \emph{Proc. Interspeech 2024}, pages 4578--4582.

\bibitem[{Choi et~al.(2025)Choi, Yeo, Chang, Watanabe, and Mortensen}]{choi2025leveraging}
Kwanghee Choi, Eunjung Yeo, Kalvin Chang, Shinji Watanabe, and David~R Mortensen. 2025.
\newblock \href {https://doi.org/10.18653/v1/2025.naacl-long.132} {Leveraging allophony in self-supervised speech models for atypical pronunciation assessment}.
\newblock In \emph{Proceedings of the 2025 Conference of the Nations of the Americas Chapter of the Association for Computational Linguistics: Human Language Technologies (Volume 1: Long Papers)}, pages 2613--2628, Albuquerque, New Mexico. Association for Computational Linguistics.

\bibitem[{Dunbar et~al.(2022)Dunbar, Hamilakis, and Dupoux}]{dunbar2022self}
Ewan Dunbar, Nicolas Hamilakis, and Emmanuel Dupoux. 2022.
\newblock Self-supervised language learning from raw audio: Lessons from the zero resource speech challenge.
\newblock \emph{IEEE Journal of Selected Topics in Signal Processing}, 16(6):1211--1226.

\bibitem[{Embick et~al.(2022)Embick, Creemers, and Goodwin~Davies}]{embick2022morphology}
David Embick, Ava Creemers, and Amy~J Goodwin~Davies. 2022.
\newblock Morphology and the mental lexicon: Three questions about decomposition.

\bibitem[{Gaskell and Marslen-Wilson(1997)}]{gaskell1997integrating}
M~Gareth Gaskell and William~D Marslen-Wilson. 1997.
\newblock Integrating form and meaning: A distributed model of speech perception.
\newblock \emph{Language and cognitive Processes}, 12(5-6):613--656.

\bibitem[{Gimenes and New(2016)}]{gimenes2016worldlex}
Manuel Gimenes and Boris New. 2016.
\newblock Worldlex: Twitter and blog word frequencies for 66 languages.
\newblock \emph{Behavior research methods}, 48:963--972.

\bibitem[{Gonnerman et~al.(2007)Gonnerman, Seidenberg, and Andersen}]{gonnerman2007graded}
Laura~M Gonnerman, Mark~S Seidenberg, and Elaine~S Andersen. 2007.
\newblock Graded semantic and phonological similarity effects in priming: evidence for a distributed connectionist approach to morphology.
\newblock \emph{Journal of experimental psychology: General}, 136(2):323.

\bibitem[{Loshchilov and Hutter(2017)}]{Loshchilov2017DecoupledWD}
Ilya Loshchilov and Frank Hutter. 2017.
\newblock \href {https://api.semanticscholar.org/CorpusID:53592270} {Decoupled weight decay regularization}.
\newblock In \emph{International Conference on Learning Representations}.

\bibitem[{Lugosch(2019)}]{lugosch2019librispeech}
Loren Lugosch. 2019.
\newblock \href {https://doi.org/10.5281/zenodo.2619474} {Librispeech alignments}.

\bibitem[{Martin et~al.(2023)Martin, Gauthier, Breiss, and Levy}]{martin2023probing}
Kinan Martin, Jon Gauthier, Canaan Breiss, and Roger Levy. 2023.
\newblock Probing self-supervised speech models for phonetic and phonemic information: a case study in aspiration.
\newblock \emph{arXiv preprint arXiv:2306.06232}.

\bibitem[{McClelland and Elman(1986)}]{mcclelland1986trace}
James~L McClelland and Jeffrey~L Elman. 1986.
\newblock The trace model of speech perception.
\newblock \emph{Cognitive psychology}, 18(1):1--86.

\bibitem[{Mikolov et~al.(2013)Mikolov, Sutskever, Chen, Corrado, and Dean}]{mikolov2013distributed}
Tomas Mikolov, Ilya Sutskever, Kai Chen, Greg~S Corrado, and Jeff Dean. 2013.
\newblock Distributed representations of words and phrases and their compositionality.
\newblock \emph{Advances in neural information processing systems}, 26.

\bibitem[{Norris(1994)}]{norris1994shortlist}
Dennis Norris. 1994.
\newblock Shortlist: A connectionist model of continuous speech recognition.
\newblock \emph{Cognition}, 52(3):189--234.

\bibitem[{Panayotov et~al.(2015)Panayotov, Chen, Povey, and Khudanpur}]{panayotov2015librispeech}
Vassil Panayotov, Guoguo Chen, Daniel Povey, and Sanjeev Khudanpur. 2015.
\newblock Librispeech: an asr corpus based on public domain audio books.
\newblock In \emph{2015 IEEE international conference on acoustics, speech and signal processing (ICASSP)}, pages 5206--5210. IEEE.

\bibitem[{Pasad et~al.(2023{\natexlab{a}})Pasad, Chien, Settle, and Livescu}]{pasad2023self}
Ankita Pasad, Chung-Ming Chien, Shane Settle, and Karen Livescu. 2023{\natexlab{a}}.
\newblock What do self-supervised speech models know about words?
\newblock \emph{arXiv preprint arXiv:2307.00162}.

\bibitem[{Pasad et~al.(2023{\natexlab{b}})Pasad, Shi, and Livescu}]{pasad2023comparative}
Ankita Pasad, Bowen Shi, and Karen Livescu. 2023{\natexlab{b}}.
\newblock Comparative layer-wise analysis of self-supervised speech models.
\newblock In \emph{ICASSP 2023-2023 IEEE International Conference on Acoustics, Speech and Signal Processing (ICASSP)}, pages 1--5. IEEE.

\bibitem[{Peterson and Lehiste(1960)}]{Peterson1960DurationOS}
Gordon~E. Peterson and Ilse Lehiste. 1960.
\newblock \href {https://api.semanticscholar.org/CorpusID:122236883} {Duration of syllable nuclei in english}.
\newblock \emph{Journal of the Acoustical Society of America}, 32:693--703.

\bibitem[{Sanabria et~al.(2023)Sanabria, Tang, and Goldwater}]{sanabria2023analyzing}
Ramon Sanabria, Hao Tang, and Sharon Goldwater. 2023.
\newblock Analyzing acoustic word embeddings from pre-trained self-supervised speech models.
\newblock In \emph{ICASSP 2023-2023 IEEE International Conference on Acoustics, Speech and Signal Processing (ICASSP)}, pages 1--5. IEEE.

\bibitem[{Zimmerman and Sapon(1958)}]{Zimmerman1958NoteOV}
Samuel~A. Zimmerman and Stanley~M. Sapon. 1958.
\newblock \href {https://api.semanticscholar.org/CorpusID:122349013} {Note on vowel duration seen cross‐linguistically}.
\newblock \emph{Journal of the Acoustical Society of America}, 30:152--153.

\end{thebibliography}

\appendix

\section{Stimulus selection}
\label{appendix:words}

We select \nunambignouns{} frequent English nouns and \nunambigverbs{} verbs which were present in the LibriSpeech dataset and were \emph{unambiguous} in their part of speech: that is, their inflected form can either be a noun or a verb, but not both.

\begin{figure}[t]
    \centering
    \includegraphics[width=0.8\linewidth]{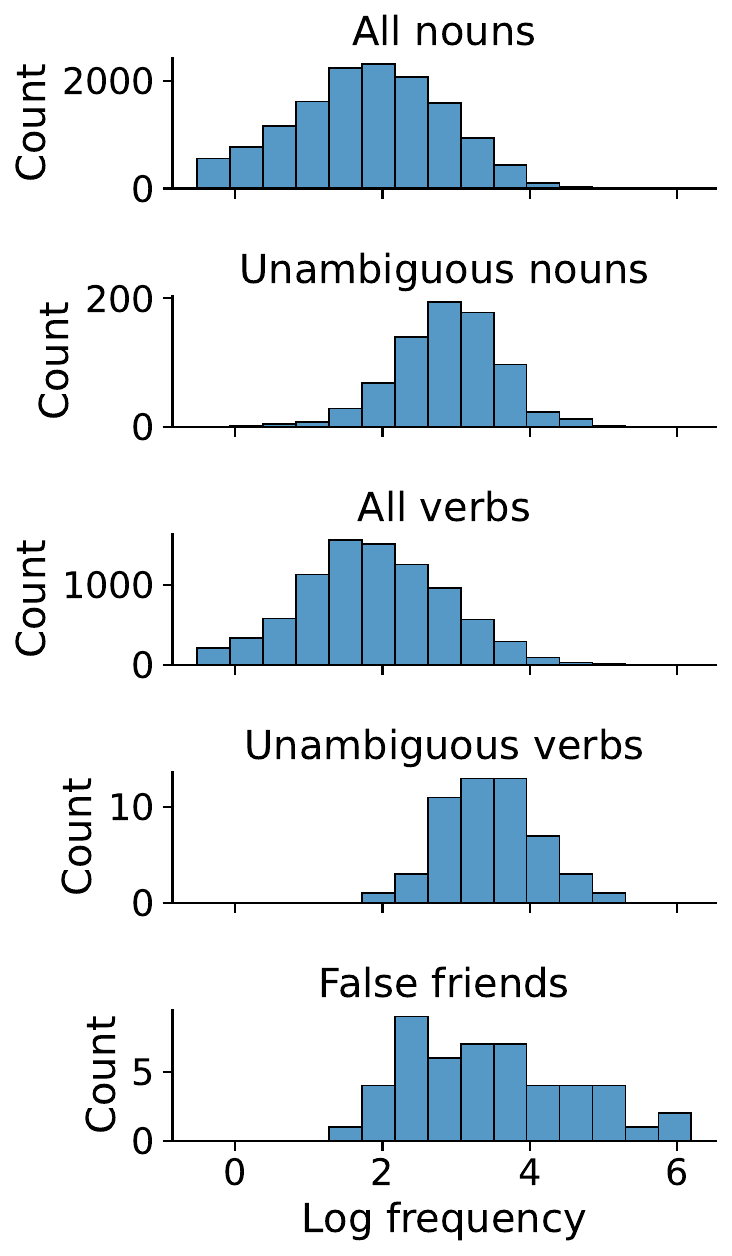}
    \caption{Frequency distributions of word stimuli used in our experiments. Each facet shows word log-10-frequency distributions on a different subset of words in the LibriSpeech corpus (\texttt{librispeech-train-clean-100} split). First two facets compare all nouns with the unambiguous nouns used in our experiments; the next two facets do the same for verbs. Final facet plots the frequency distributions of false friend items, used in \Cref{sec:exp-false-friends}.}
    \label{fig:word-frequencies}
\end{figure}

Specifically, for each word type in the dataset, we calculated the distribution of its part-of-speech labels in the corpus, tagged with spaCy's \texttt{en_core_web_trf} part-of-speech tagger. We retained only those noun types which had no attested verb instances, and vice versa. We then manually excluded residual ambiguous items from the resulting list. \Cref{fig:word-frequencies} shows the frequency distribution of the retained nouns and verbs and compares it with the marginal noun/verb frequency distribution in the corpus. Our stimulus filtering procedure does not select for long-tail rare words on which the models may have poorly calibrated representations --- it seems to do just the opposite, selecting a relatively high-frequency subsample of word types.

\Cref{tbl:stimuli-nouns,tbl:stimuli-verbs} give the complete list of unambiguous nouns and verbs used in our experiments.

\begin{table*}[t]
\centering
\caption{Unambiguous nouns (\nunambignouns{}) used in our experiments. Continued in \Cref{tbl:stimuli-nouns-2}.}
\label{tbl:stimuli-nouns}
\resizebox{0.9\linewidth}{!}{
\begin{tabular}{lllllllll}
\hline
ability & accident & accomplishment & achievement & acquaintance & acre & action & activity & actor \\
adder & advantage & adventure & advertiser & affair & affection & age & agent & agony \\
allusion & ambition & ancient & angel & angle & animal & ankle & apartment & ape \\
apple & application & apprehension & argument & army & arrangement & arrow & artery & article \\
artist & ash & assistant & association & attention & attitude & aunt & author & authority \\
baby & bag & baker & ball & band & bank & banker & banner & barn \\
barrel & barricade & barrier & basket & beast & beauty & bed & bee & beech \\
beggar & being & bell & bench & bird & biscuit & blade & blanket & boat \\
body & bond & bone & book & bottle & bough & boundary & boy & branch \\
breast & brother & brow & buccaneer & bud & buffalo & bull & bullet & bundle \\
burden & button & cabin & cage & cake & candle & cannon & canoe & canyon \\
captain & captive & car & card & carriage & cart & case & castle & cat \\
cave & cedar & cell & cent & century & ceremony & chair & chamber & champion \\
channel & chapter & characteristic & cheek & cheese & chicken & chief & chimney & church \\
circumstance & citizen & city & clerk & cliff & cloak & clock & cloud & club \\
coal & coat & coin & colony & column & combination & commander & commissioner & common \\
community & companion & company & competitor & complaint & composition & comrade & conception & conclusion \\
condition & connection & consideration & contemporary & contribution & conversation & cord & cordial & corn \\
corpse & cottage & counsel & country & couple & course & cousin & crane & creature \\
crime & criminal & crystal & cup & current & curtain & custom & customer & danger \\
daughter & day & death & debt & deck & degree & demonstration & depth & description \\
desk & detail & device & devil & diamond & difficulty & dinner & direction & disappointment \\
disaster & discovery & disease & doctor & doctrine & dog & dollar & door & doorway \\
dozen & drawer & duty & eagle & ear & eel & effort & egg & elbow \\
elder & element & emotion & enchantment & enemy & energy & engagement & engine & enterprise \\
error & estate & evening & event & evil & example & exception & exclamation & excursion \\
exertion & expectation & expense & experiment & expert & explanation & expression & extremity & fact \\
factor & factory & failure & fairy & family & farmer & father & feather & feature \\
fellow & female & field & finger & fir & fist & flag & fleet & flight \\
flood & floor & flower & folk & forest & fort & fortune & foundation & fountain \\
fowl & fragment & friend & frog & fruit & fund & fur & gale & game \\
garden & garment & gate & general & generation & gesture & ghost & giant & gift \\
girl & glacier & glass & glimpse & god & good & government & gown & grain \\
grape & grave & grove & guest & guinea & gun & habit & hair & hall \\
hardship & hat & heart & heaven & hedge & heel & height & hen & hero \\
hill & hip & historian & history & hole & holiday & home & horn & horror \\
horse & host & hotel & hour & house & humor & hundred & hunter & husband \\
hut & idea & ideal & illusion & image & imagination & impression & improvement & impulse \\
inch & incident & income & inconvenience & indication & individual & injury & insect & instinct \\
institution & instruction & instrument & intention & interruption & interval & investigation & invitation & island \\
jackal & jaw & jewel & joy & junior & justice & key & kind & king \\
kingdom & knee & knight & laborer & lad & lady & lake & lamp & lane \\
language & lantern & law & lawyer & leader & league & leg & lesson & letter \\
liberty & lily & limb & lion & lip & list & lord & loss & lot \\
lover & luxury & machine & magistrate & maiden & maker & manner & map & maple \\
marriage & martian & mast & material & maxim & meadow & meal & medicine & melody \\
member & memory & merchant & message & messenger & meter & method & mile & mill \\
million & mink & minute & miracle & mirror & misery & misfortune & mist & mode \\
model & moment & monk & monkey & monster & month & monument & mood & moral \\
mortal & mosquito & motion & motive & mountain & movement & mule & multitude & murderer \\
muscle & musket & musketeer & muskrat & myriad & mystery & nation & native & nature \\
necessity & neck & needle & neighbor & nerve & net & newspaper & niece & night \\
noble & noise & nose & notion & novel & nut & oar & oath & objection \\
obligation & observation & obstacle & occupation & odor & office & officer & official & one \\
operation & opinion & opponent & opportunity & orchard & organ & orphan & other & outline \\
owner & oyster & pace & page & pair & palace & paper & parent & parish \\
park & particular & partner & party & passage & passion & patch & path & patient \\
peak & pearl & peasant & peculiarity & people & perception & performance & period & person \\
petticoat & philosopher & physician & pig & pillar & pillow & pine & pipe & pirate \\
pistol & pit & plain & plane & planet & plank & plantation & plate & platform \\
plum & poet & pole & politician & pool & portrait & position & possession & possibility \\
post & potato & power & prayer & precaution & prejudice & preparation & price & priest \\
prince & princess & principle & prisoner & privilege & problem & product & professor & prophet \\
proportion & proposal & proposition & prospect & provision & pupil & purpose & pursuit & quality \\
quantity & quarter & rascal & ray & reader & reality & recollection & regiment & region \\
relation & relative & religion & remnant & representative & resolution & resource & responsibility & ribbon \\
rider & right & river & road & robber & robe & roof & room & rope \\
ruler & rumor & sailor & saint & savage & scene & scheme & scholar & science \\
scrap & scripture & sea & secret & section & senior & sensation & sentiment & sentinel \\
servant & service & sex & shadow & shaft & sheet & shilling & shirt & shoe \\
shop & shore & signature & silk & singer & sister & situation & skin & sky \\
sledge & sleeve & slice & slope & snake & soldier & son & song & sorrow \\
soul & space & spear & specimen & spectacle & spectator & speculation & speech & spirit \\
stable & stage & stair & star & statement & station & steamer & stone & story \\
stranger & street & string & stroke & structure & student & success & suggestion & summit \\
suspicion & sword & sympathy & system & table & tale & talent & tank & task \\
teacher & temple & temptation & tendency & tent & terror & theory & thing & thousand \\
threat & tiger & title & toe & tongue & torrent & tower & town & trader \\
\hline
\end{tabular}
}
\end{table*}

\begin{table*}[t]
\centering
\caption{Unambiguous nouns (\nunambignouns{}) used in our experiments. Continuation of \Cref{tbl:stimuli-nouns}.}
\label{tbl:stimuli-nouns-2}
\resizebox{0.9\linewidth}{!}{
\begin{tabular}{lllllllll}
\hline
traveler & tree & trial & tribe & troop & truth & turkey & turnip & turtle \\
twig & twin & type & tyrant & valley & vapor & variety & vein & verse \\
vessel & vice & victim & victory & village & villain & vine & virtue & vision \\
visitor & voice & volume & wagon & wall & wand & war & warrior & way \\
weapon & week & well & wheel & window & wine & wing & wit & wood \\
word & worker & world & writer & yard & year &  &  &  \\
\hline
\end{tabular}
}
\end{table*}

\begin{table*}[t]
\centering
\caption{Unambiguous verbs (\nunambigverbs{}) used in our experiments.}
\label{tbl:stimuli-verbs}
\begin{tabular}{llllllll}
\hline
allow & appear & arise & ask & attract & become & begin & belong \\
bring & carry & come & contain & continue & depend & describe & deserve \\
do & eat & enter & exist & extend & follow & forward & give \\
grow & happen & hear & insist & involve & learn & occur & owe \\
own & perceive & possess & prevent & prove & put & receive & remember \\
remind & require & send & serve & shine & sit & speak & suggest \\
take & tell & tend & think &  &  &  &  \\
\hline
\end{tabular}
\end{table*}

\section{Supplementary results}

\subsection{Extended results of main analyses}
\label{sec:extended-results}

\begin{figure}
    \includegraphics[width=\linewidth]{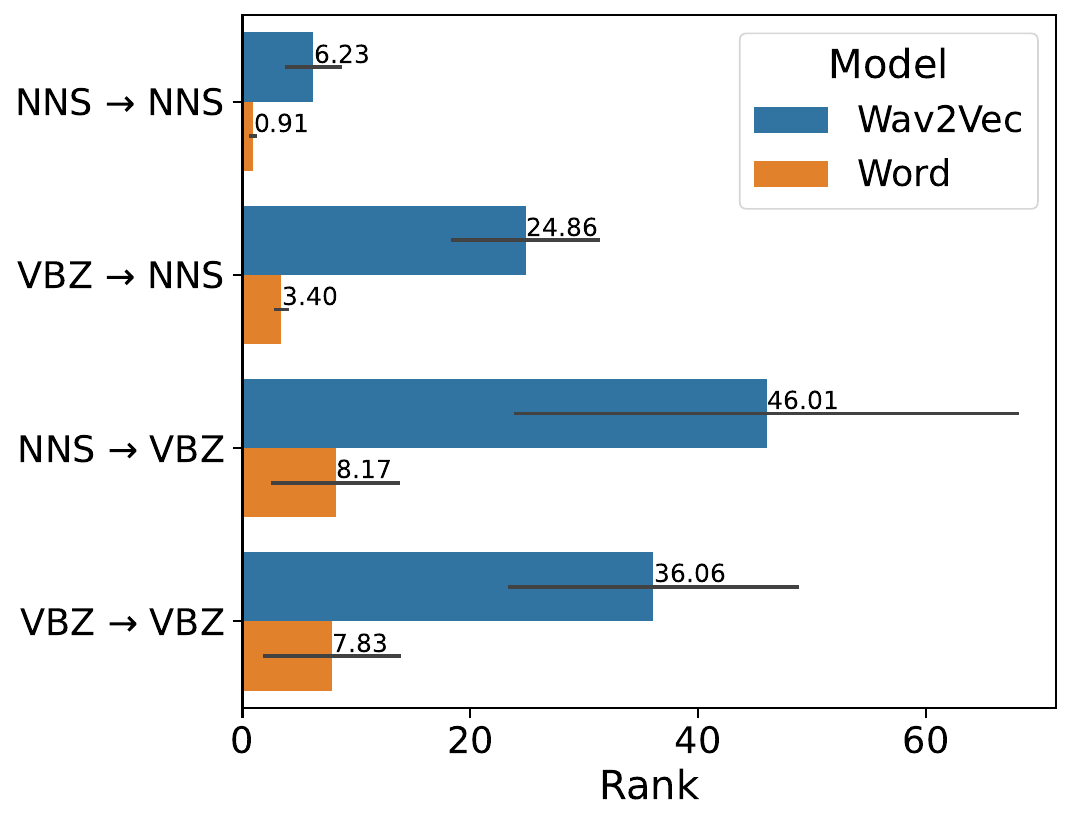}
    \caption{Rank mean and standard error estimates for the main morphology analysis of \Cref{sec:exp-morphology,fig:transfer-morph}. Mean values differ slightly from \cref{fig:transfer-morph} because this figure estimates performance pooled within single target word pairs in order to derive a meaningful uncertainty measure, whereas \cref{fig:transfer-morph} pools all trials together regardless of target word pair.}
    \label{fig:nnvb-barplot}
\end{figure}
\begin{figure*}
    \centering
    \includegraphics[width=0.8\linewidth]{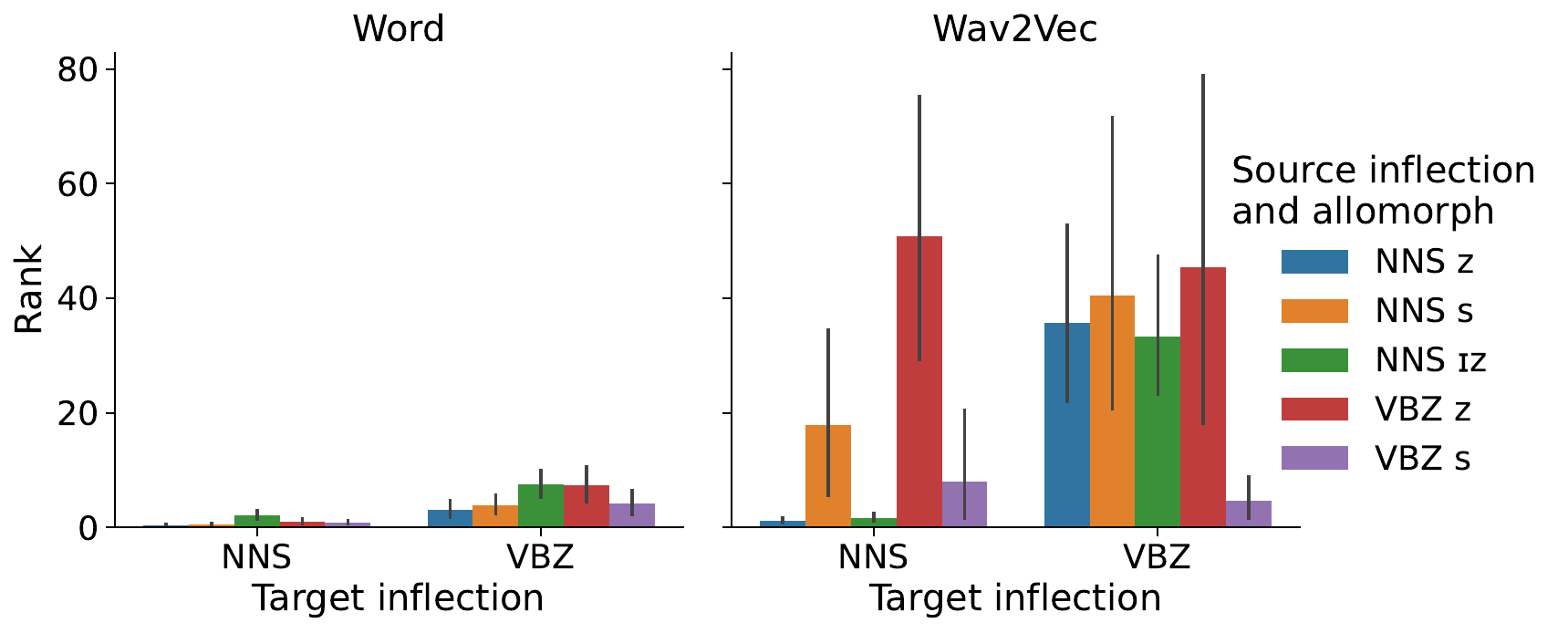}
    \caption{Rank mean and standard error estimates decomposed by both the morpheme and particular allomorph expressed in the source word pair.}
    \label{fig:nnvb-allomorph-barplot}
\end{figure*}
\begin{figure*}
    \centering
    \includegraphics[width=\linewidth]{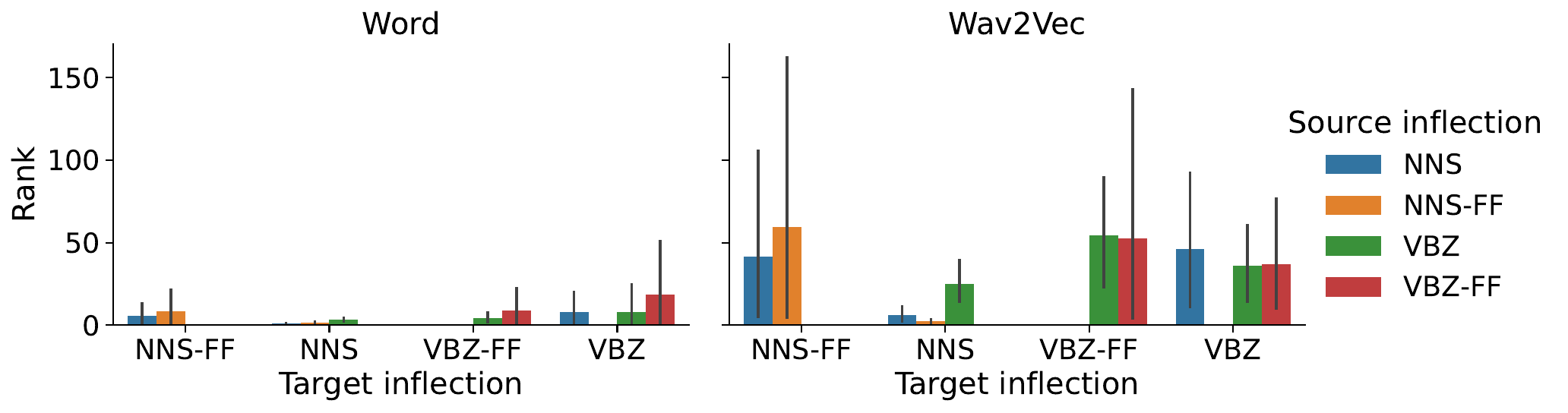}
    \caption{Rank mean and standard error estimates for the false friend analysis of \Cref{sec:exp-false-friends,fig:false-friends}.}
    \label{fig:ff-barplot}
\end{figure*}
\begin{figure*}
    \centering
    \includegraphics[width=0.7\linewidth]{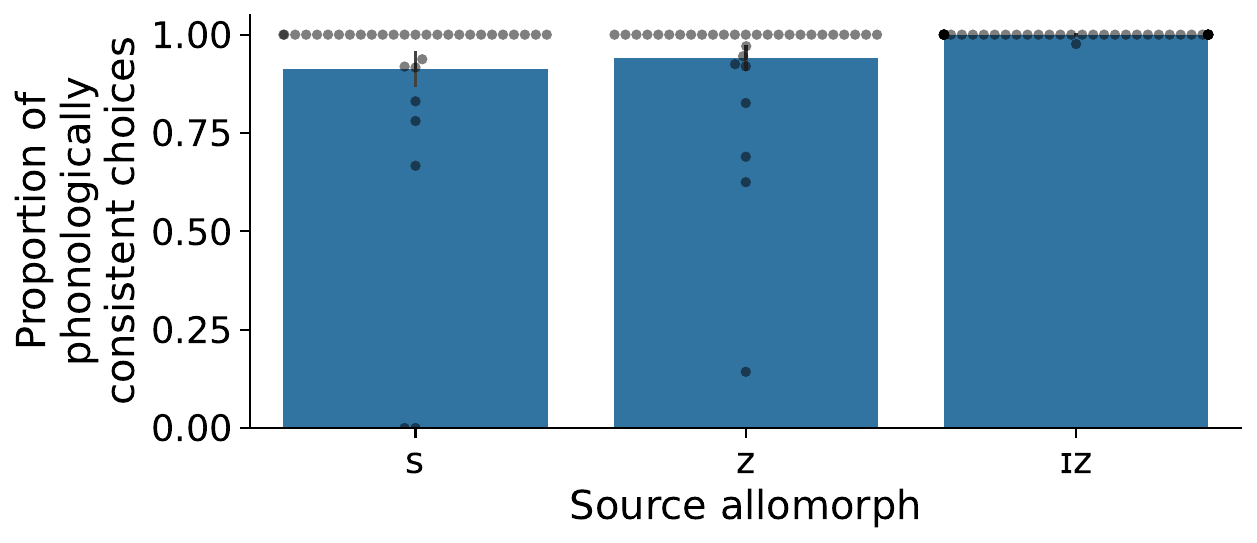}
    \caption{Proportions of phonologically consistent choices in the forced-choice experiment of \Cref{sec:exp-forced-choice}. Each point corresponds to a forced-choice triple; results are grouped by the particular sound present in the source word pair.}
    \label{fig:fc-barplot}
\end{figure*}
\Cref{fig:nnvb-barplot,fig:nnvb-allomorph-barplot,fig:ff-barplot,fig:fc-barplot} offer more statistically detailed versions of the main plots \Cref{fig:transfer-morph,fig:transfer-allomorph,fig:false-friends,fig:forced-choice}, respectively.

\begin{figure*}
    \centering
    \includegraphics[width=\linewidth]{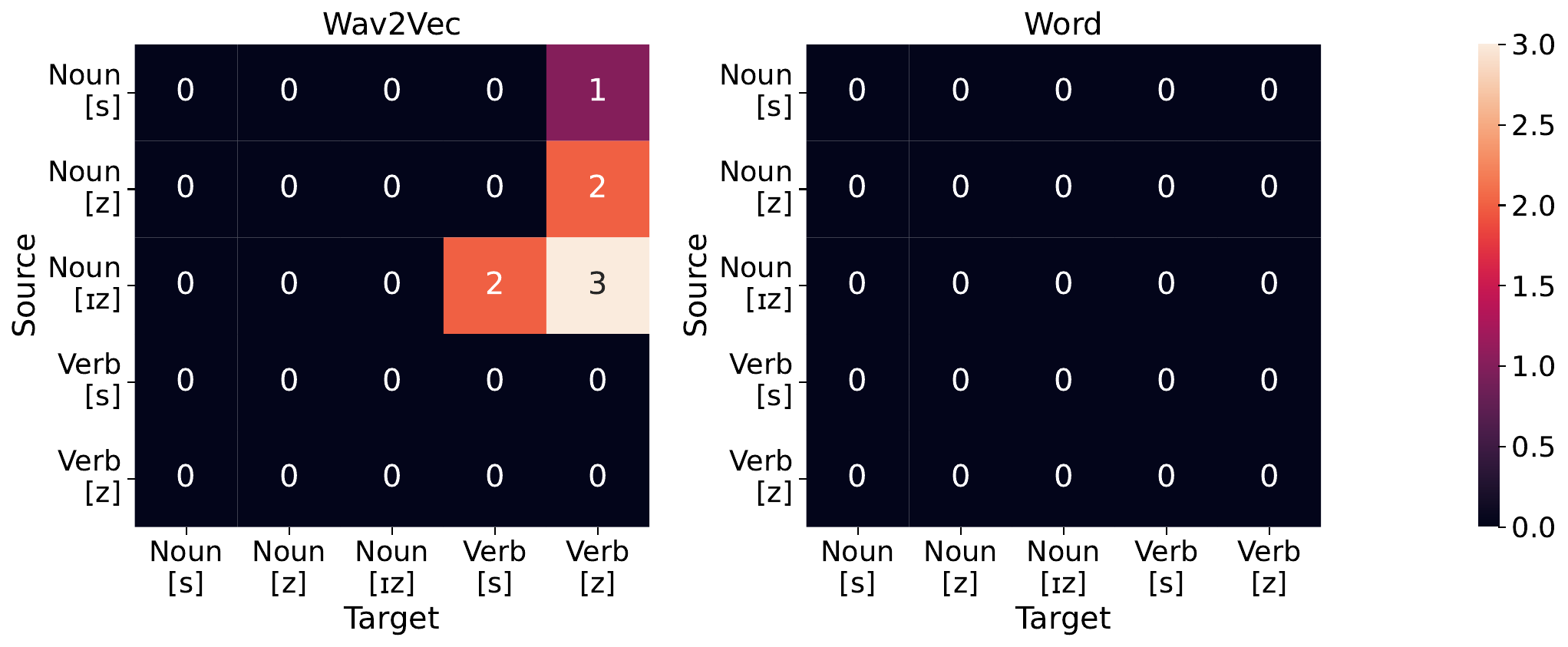}
    \caption{Median rank results for analogy within and between allomorphs of noun and verb inflections. Compare with mean values in \Cref{fig:transfer-allomorph}.}
    \label{fig:transfer-allomorph-median}
\end{figure*}
\begin{figure*}
    \centering
    \includegraphics[width=0.9\linewidth]{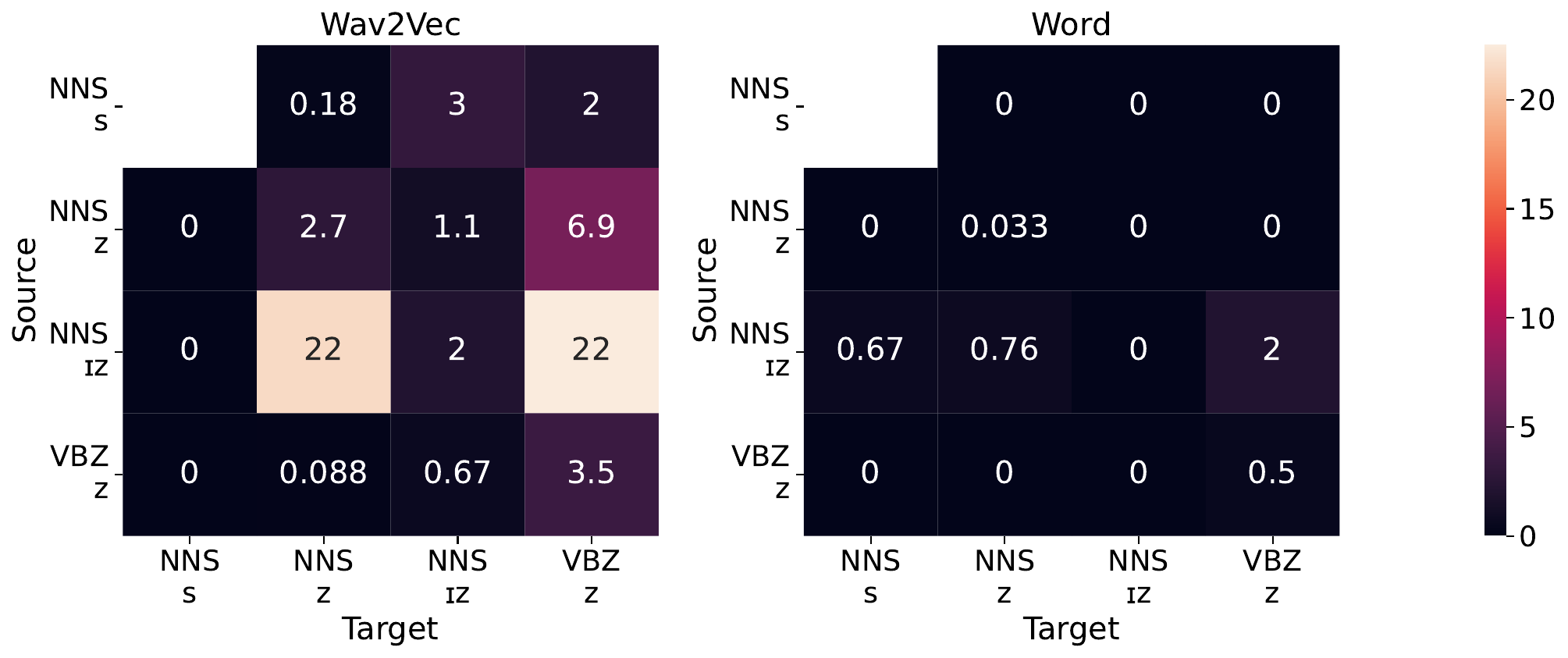}
    \caption{Analogy within and between allomorphs of noun and verb inflections, estimated on the data split \texttt{librispeech-test-clean} (see \Cref{sec:exp-morphology-heldout}). Heatmaps show an average rank metric (0 is best; random guessing is \randomChanceRank). The word probe in the right panel exhibits strongly reduced sensitivity (i.e., improved performance) to allomorphic contrasts. Analogous to \Cref{fig:transfer-allomorph}.}
    \label{fig:transfer-allomorph-heldout}
\end{figure*}
\begin{figure*}
    \centering
    \includegraphics[width=\linewidth]{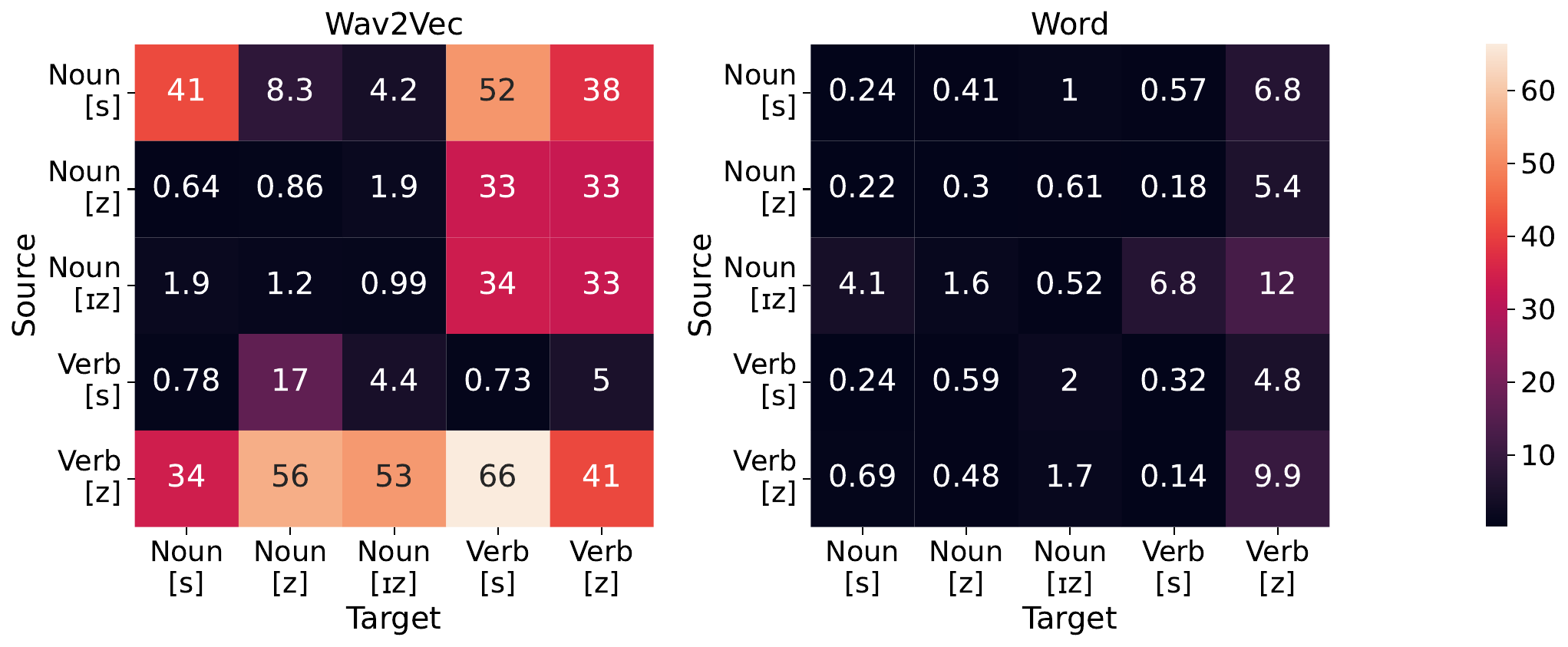}
    \caption{Phoneme-pooled analogy results within and between allomorphs of noun and verb inflections, estimated without word-level pooling (\cref{sec:phoneme-pooling}). Analogous to \Cref{fig:transfer-allomorph}.}
    \label{fig:transfer-allomorph-pc}
\end{figure*}

\begin{figure}
    \centering
    \includegraphics[width=\linewidth]{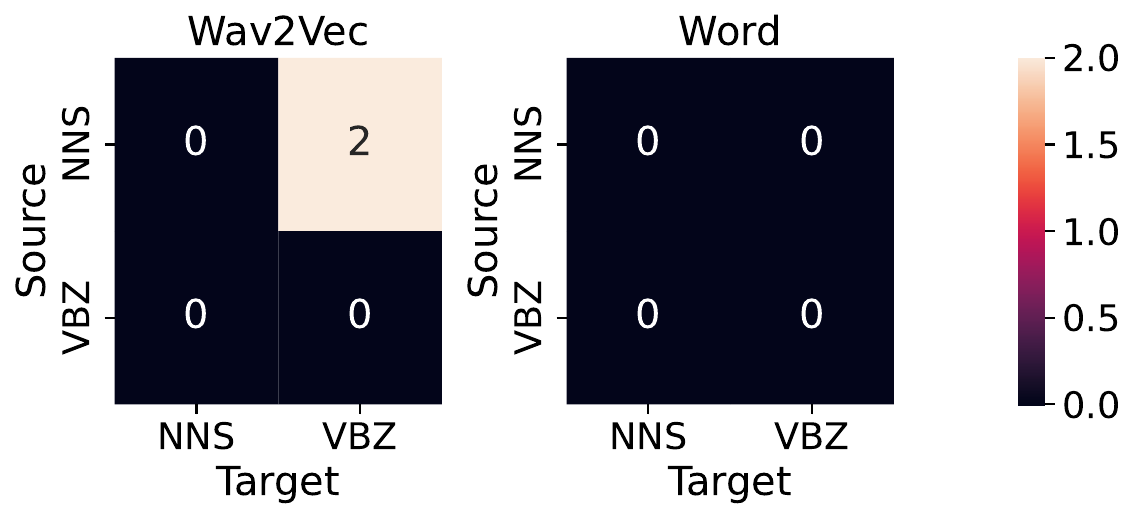}
    \caption{Median rank results for analogy within and between noun/verb inflections on the Wav2Vec baseline and word probe. Compare with mean values in \Cref{fig:transfer-morph}.}
    \label{fig:transfer-morph-median}
\end{figure}
\begin{figure}
    \centering
    \includegraphics[width=\linewidth]{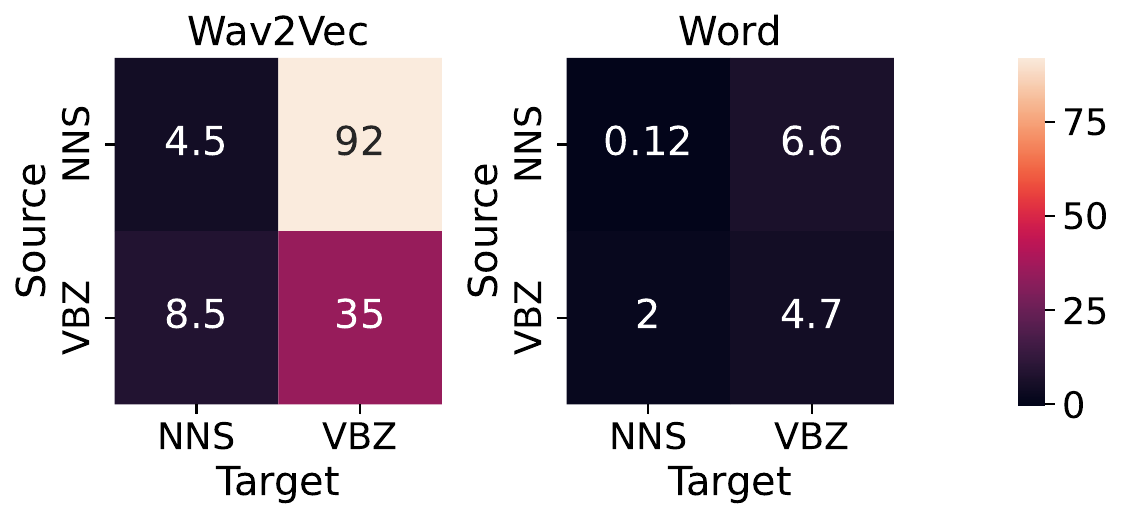}
    \caption{Mean rank values (lower is better) from analogy within and between noun/verb inflections on the Wav2Vec baseline and word probe, estimated on the data split \texttt{librispeech-test-clean} (see \Cref{sec:exp-morphology-heldout}). Analogous to \Cref{fig:transfer-morph}.}
    \label{fig:transfer-morph-heldout}
\end{figure}
\begin{figure}
    \centering
    \includegraphics[width=\linewidth]{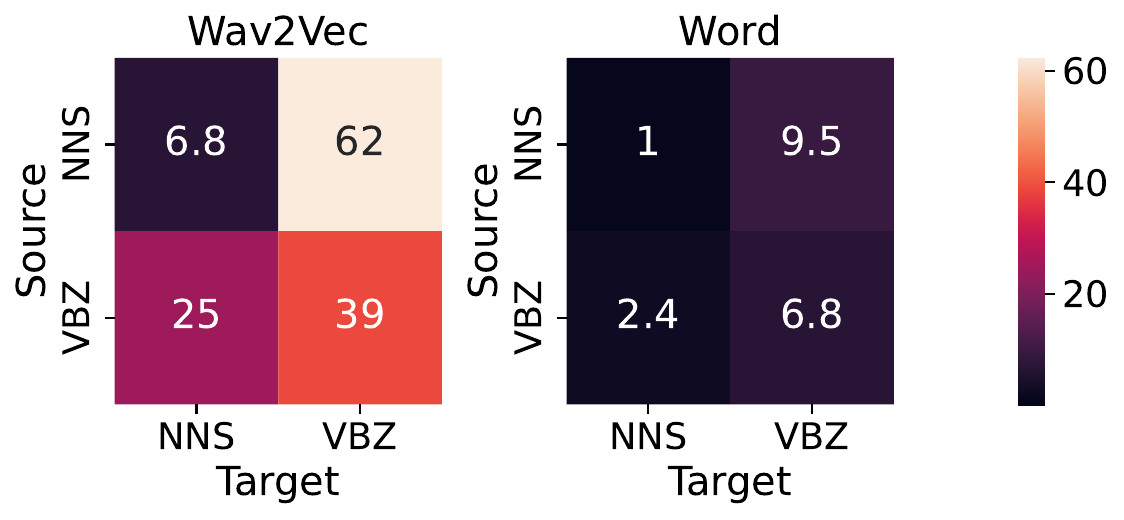}
    \caption{Phoneme-pooled analogy results within and between noun/verb inflection on the Wav2Vec baseline and word probe, estimated without word level pooling (\cref{sec:phoneme-pooling}). Analogous to \Cref{fig:transfer-morph}.}
    \label{fig:transfer-morph-pc}
\end{figure}

\begin{figure}
    \centering
    \includegraphics[width=\linewidth]{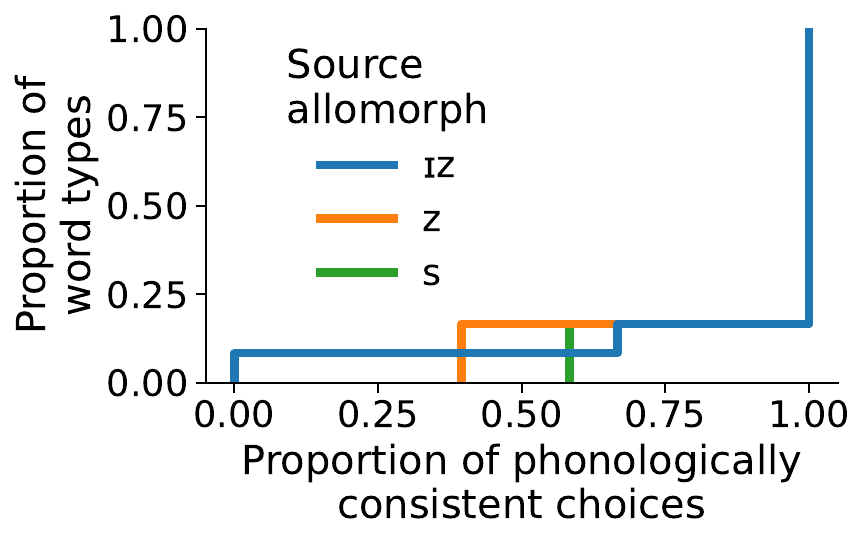}
    \caption{Cumulative distribution of preferences for the forced-choice item obeying distributional constraints, estimated on the data split \texttt{librispeech-test-clean} (see \Cref{sec:exp-morphology-heldout}). Analogous to \Cref{fig:forced-choice}.}
    \label{fig:forced-choice-heldout}
\end{figure}
\begin{figure}
    \centering
    \includegraphics[width=\linewidth]{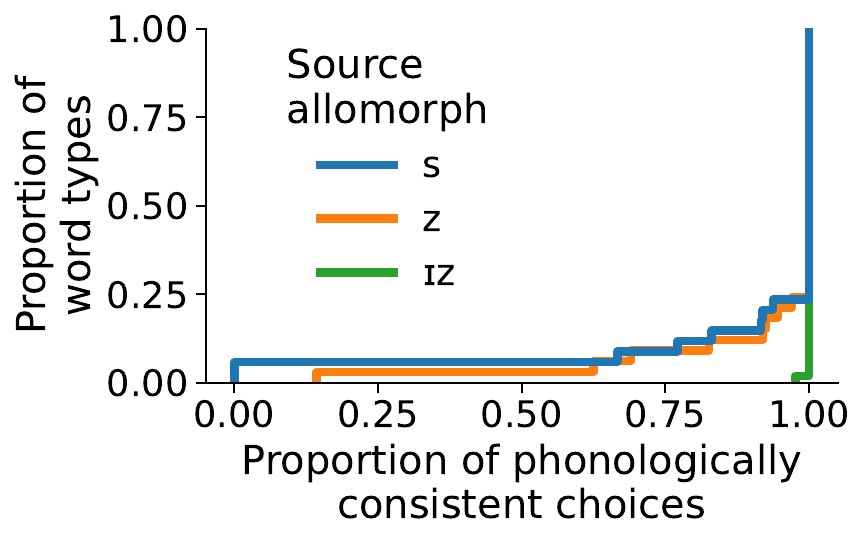}
    \caption{Cumulative distribution of preferences for the forced-choice item obeying distributional constraints, estimated without word-level pooling (\Cref{sec:phoneme-pooling}). Analogous to \Cref{fig:forced-choice}.}
    \label{fig:forced-choice-pc}
\end{figure}

Median rank values are given for the equivalent mean rank visualizations in the main text: \Cref{fig:transfer-morph-median} (analogous to \Cref{fig:transfer-morph}) and \Cref{fig:transfer-allomorph-median} (analogous to \Cref{fig:transfer-allomorph}).

\subsubsection{Regression results}
\label{appendix:regression}
\begin{table*}
\caption{Full regression model fits for predicting rank outcomes from individual trial properties, following the regression model given in \Cref{eqn:regression}.}
\label{tbl:regression-results}
\resizebox{\linewidth}{!}{
\begin{tabular}{lrr}
\toprule
Variable & Wav2Vec & Word \\
\midrule
\texttt{Intercept} & 0.53 & 0.10 \\
\texttt{allomorph_from=S × allomorph_to=S} & -0.20 & -0.08 \\
\texttt{allomorph_from=S} & -0.06 & 0.06 \\
\texttt{allomorph_to=S} & -0.30 & 0.06 \\
\texttt{from_freq} & -0.38 & -0.04 \\
\texttt{inflection_from=VBZ × allomorph_from=S × allomorph_to=S} & -0.77 & -0.01 \\
\texttt{inflection_from=VBZ × allomorph_from=S} & 0.46 & -0.01 \\
\texttt{inflection_from=VBZ × allomorph_to=S} & 0.87 & -0.03 \\
\texttt{inflection_from=VBZ × inflection_to=VBZ × allomorph_from=S × allomorph_to=S} & 0.46 & -0.31 \\
\texttt{inflection_from=VBZ × inflection_to=VBZ × allomorph_from=S} & 1.41 & 0.25 \\
\texttt{inflection_from=VBZ × inflection_to=VBZ × allomorph_to=S} & -0.71 & 0.07 \\
\texttt{inflection_from=VBZ × inflection_to=VBZ} & -5.17 & -0.13 \\
\texttt{inflection_from=VBZ} & 0.07 & 0.14 \\
\texttt{inflection_to=VBZ × allomorph_from=S × allomorph_to=S} & 0.48 & 0.06 \\
\texttt{inflection_to=VBZ × allomorph_from=S} & -2.06 & 0.08 \\
\texttt{inflection_to=VBZ × allomorph_to=S} & -1.57 & -0.21 \\
\texttt{inflection_to=VBZ} & 7.53 & 0.05 \\
\texttt{to_freq} & -1.05 & 0.01 \\
\bottomrule
\end{tabular}
}
\end{table*}
\Cref{tbl:regression-results} gives the full regression models estimated independently on the rank outcomes from the Wav2Vec and Word models. The interaction strength values of \Cref{fig:interaction-strength} are generated by comparing the difference of absolute values of all rows of this table which are interactions (i.e., terms including ``×'').

\subsection{Results on held-out LibriSpeech data}
\label{sec:exp-morphology-heldout}

The results reported in the main text are evaluated on the \texttt{librispeech-train-clean-100} data split of the LibriSpeech corpus, which was seen both during the pretraining of the Wav2Vec model and the word probe. To validate these findings, we also evaluated on the held-out split \texttt{librispeech-test-clean} set, which was not used during training or hyperparameter selection of either Wav2Vec or the word probe.

However, this test set is relatively small ($\sim 5$ hours), leading to substantial data sparsity in our more granular analyses, such as the allomorphy evaluation of \Cref{sec:exp-allomorph}. For example, when we restrict our analysis to word types appearing in the \texttt{test-clean} split at least 5 times, there is exactly one regular unambiguous English noun plural usable in our analysis: \emph{hearts}.

For this reason, the main text reports results on the much larger \texttt{train-clean-100} set, where we can evaluate these trends of interest. This section shows that results on the held-out \texttt{test-clean} set qualitatively---and, where measurable, quantitatively---replicate those reported in the main text. \Cref{fig:transfer-morph-heldout} provides an analogous plot to \Cref{fig:transfer-morph}, \Cref{fig:transfer-allomorph-heldout} to \Cref{fig:transfer-allomorph}, and \Cref{fig:forced-choice-heldout} to \Cref{fig:forced-choice}. Some of the cells of \Cref{fig:transfer-allomorph-heldout} are null due to data sparsity.

The quantitative trends given in the main text hold in the held-out dataset:
\begin{itemize}
    \item Wav2Vec performs worse than the word probe on the analogy text ($t\approx 2.00, p \approx 0.0481$).
    \item Nouns are better predicted than verbs in analogies on the word probe representations ($t \approx -2.89, p \approx 0.00584$)
\end{itemize}

\subsection{Results without word-level pooling}
\label{sec:phoneme-pooling}

It is possible that the evaluation as given in the main text may disadvantage Wav2Vec relative to the word probe. Our word probe constrains all frames across the span of a word to converge to a single type-level representation, potentially erasing dynamic temporal information within Wav2Vec's frame representations (\Cref{sec:probe-model}). This kind of code might be less sensitive to the average-pooling embedding method of \Cref{sec:word-embeddings} compared to the Wav2Vec model. It is possible, then, that our analogy evaluations are unfairly biased toward the word probe.

To address this, we introduce a \emph{phoneme-level pooling} version of the experiments in the main text. In this setting, rather than submitting word embeddings (\cref{eqn:wordrep}) to vector arithmetic (\cref{eqn:analogy}), we operate on individual \emph{phoneme embeddings} from critical points within words.

For each regularly inflected word (e.g. \emph{shirts}), we define a \emph{constancy point} as the final phoneme it shares with its base form (i.e., the /t/ of \emph{shirts}). Let $f_c(w_i)$ be the mean-pooled representation of the audio frames spanning the constancy point, and $f_f(w_i)$ be the mean-pooled representation of the final phoneme in the word (here /s/). Our difference vectors now compute the relationship $f_f(w_i) - f_c(w_i)$: the movement of the model representation from a point prior to the inflection to a point after the inflection. 

To solve an analogy \emph{shirt} : \emph{shirts} :: \emph{cheese} : \underline{\hspace{1cm}}, we perform an updated version of \Cref{eqn:analogy} for some token $i$ of \emph{shirt} and some token $j$ of \emph{cheese}:
\begin{equation}
    \hat d = f_f(\text{shirts}_i) - f_c(\text{shirts}_i) + f_c(\text{cheese}_j)
\end{equation}

We qualitatively and quantitatively replicate the outcomes shown in the main text under this evaluation. Analogs of the main result figures are given in \Cref{fig:transfer-morph-pc} (analogous to \Cref{fig:transfer-morph}), \Cref{fig:transfer-allomorph-pc} (analogous to \Cref{fig:transfer-allomorph}), and \Cref{fig:forced-choice-pc} (analogous to \Cref{fig:forced-choice}). This confirms that the word probe meaningfully alters the model's dynamic response to individual sounds, rather than simply generating a response which is more amenable to word-level average-pooling.

\subsection{Same-word evaluation}
\label{sec:same-word}

Our analogy method asks whether a predicted word embedding $\hat d$ (\Cref{eqn:analogy}) is close to any token of a desired word. For example, on the analogy \emph{shirt} : \emph{shirts} :: \emph{cheese} : \underline{\hspace{1cm}}, we compute the minimum distance between the predicted $\hat d$ and any token of the desired word \emph{cheeses}. This method can effectively capture the relationship between base and inflected forms of nouns and verbs in our main evaluations.

\begin{table}
    \centering
    \begin{tabular}{l|ll}
    \toprule
    Model & Noun & Verb \\
    \midrule
    Wav2Vec & 10.59 & 1.22 \\
    Word probe & 1.94 & 0.135 \\
    \bottomrule
    \end{tabular}
    \caption{Mean rank outcomes of the same-word evaluation of \Cref{sec:same-word}. These values are a soft lower bound for the results in \Cref{fig:transfer-morph}.}
    \label{tbl:same-word}
\end{table}
\begin{figure}
    \centering
    \includegraphics[width=\linewidth]{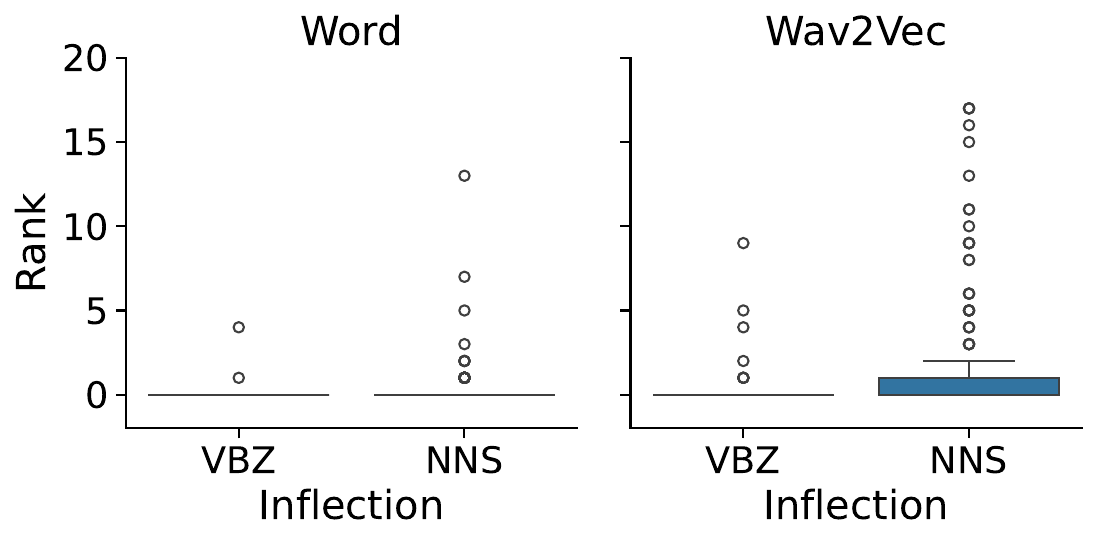}
    \caption{Rank outcomes of the same-word evaluation of \Cref{sec:same-word}. Our analogy method can be applied to map between distinct tokens of single word types; this establishes a performance bound for the results of our main experiments in \Cref{sec:experiments}.}
    \label{fig:same-word}
\end{figure}

In a supplementary evaluation we asked whether this same logic applies across tokens of a single word type. If this were true, it would demonstrate that word types are \textbf{coherent} in embedding space, and establish a soft lower bound (maximal performance) for our transfer evaluations in the main text. We draw embeddings of arbitrary bases and their inflected forms $x_{\text{self}}$ (e.g. \emph{shirt}) and $y_{\text{self}}$ (\emph{shirts}), and split these into two disjoint ``source'' and ``target'' sets. We then ask whether difference vectors computed on embeddings from the source set can be transferred to embeddings from the target set:
\begin{equation}
    \hat d_{\text{self}} = x^{\text{source}}_{\text{self}} - a^{\text{source}}_{\text{self}} + c^{\text{target}}_{\text{self}}
\end{equation}
We compare the predicted $\hat d_{\text{self}}$ to the unseen target embeddings $d_{\text{self}}^{\text{target}}$ and compute the same mean rank metric as described in \Cref{sec:experiments}. Mean rank outcomes are given in \Cref{tbl:same-word} for the Wav2Vec baseline and the word probe; \Cref{fig:same-word} compares outcomes of the same-word evaluation within nouns and verbs for the two models.

By comparing these values to the main results in \Cref{fig:transfer-morph}, we see that the analogy results across words in some cases reach the upper bound established by the same-word evaluation in both models. For example, the word probe performs best in noun--noun analogies with a mean rank of 1; slightly exceeding the model's same-word result of 1.9.

\subsection{Model optimization}
\label{appendix:training}

\begin{table}[th]
    \centering
    \begin{tabular}{l|l}
    \toprule
    Hyperparameter & Value \\
    \midrule
    Dimensionality & 32 \\
    Margin $m$ & 0.37590 \\
    \midrule
    Learning rate & 0.00108 \\
    Weight decay & 0.00607 \\
    \bottomrule
    \end{tabular}
    \caption{Optimal hyperparameters for the word probe at the 8th layer of Wav2Vec Base, selected on a held-out development set in order to maximize discriminability of word categories.}
    \label{tbl:hyperparameters}
\end{table}
The word probe model computes a 32-dimensional representation of each frame through a linear transform of Wav2Vec's 768-dimensional activation (\Cref{eqn:wav2vec-frame}), with $32\times 768=24,576$ total learnable parameters.
We estimate individual probes for each layer of the Wav2Vec base model with AdamW \citep{Loshchilov2017DecoupledWD}, minimizing the hinge loss given in \Cref{eqn:loss} and early-stopping using a held-out validation dataset. Model hyperparameters (the margin parameter $m$, learning rate, and weight decay) are selected using a separate validation set in order to maximize the mean average precision of a classifier mapping from probe frames (\Cref{eqn:wav2vec-frame}) to word categories, following the word-annotated LibriSpeech corpus. Optimal values for the model analyzed in the main results of this paper (mapping from layer 8 of Wav2Vec) are given in \Cref{tbl:hyperparameters}.

A complete hyperparameter search for a single layer requires approximately 12 hours on a single NVIDIA TITAN V GPU.

\subsection{Forced-choice experiment}
\label{sec:fc-length-confound}

\Cref{tbl:fc-materials} gives the full set of forced-choice triples used in \Cref{sec:exp-forced-choice}.
\begin{table*}[ht]
    \centering
    \resizebox{\linewidth}{!}{
    \begin{tabular}{ll|llll}
        \toprule
        Base & Base (IPA) & Consistent & Consistent (IPA) & Inconsistent & Inconsistent (IPA) \\
        \midrule
\emph{bay} & /\textipa{beI}/ & \emph{bays} & /\textipa{beIz}/ & \emph{base} & /\textipa{beIs}/ \\
\emph{decree} & /\textipa{dIkri}/ & \emph{decrees} & /\textipa{dIkriz}/ & \emph{decrease} & /\textipa{dIkris}/ \\
\emph{den} & /\textipa{dEn}/ & \emph{dens} & /\textipa{dEnz}/ & \emph{dense} & /\textipa{dEns}/ \\
\emph{dew} & /\textipa{du}/ & \emph{dews}, \emph{dues} & /\textipa{duz}/ & \emph{deuce} & /\textipa{dus}/ \\
\emph{display} & /\textipa{dIspleI}/ & \emph{displays} & /\textipa{dIspleIz}/ & \emph{displace} & /\textipa{dIspleIs}/ \\
\emph{fall} & /\textipa{fOl}/ & \emph{falls} & /\textipa{fOlz}/ & \emph{false} & /\textipa{fOls}/ \\
\emph{fay} & /\textipa{feI}/ & \emph{phase} & /\textipa{feIz}/ & \emph{face} & /\textipa{feIs}/ \\
\emph{fear} & /\textipa{fir}/ & \emph{fears} & /\textipa{firz}/ & \emph{fierce} & /\textipa{firs}/ \\
\emph{flee} & /\textipa{fli}/ & \emph{flees} & /\textipa{fliz}/ & \emph{fleece} & /\textipa{flis}/ \\
\emph{for} & /\textipa{fOr}/ & \emph{fours} & /\textipa{fOrz}/ & \emph{force} & /\textipa{fOrs}/ \\
\emph{gray} & /\textipa{greI}/ & \emph{gray's}, \emph{grays}, \emph{graze} & /\textipa{greIz}/ & \emph{grace} & /\textipa{greIs}/ \\
\emph{hahn} & /\textipa{hAn}/ & \emph{hahn's} & /\textipa{hAnz}/ & \emph{hans} & /\textipa{hAns}/ \\
\emph{hen} & /\textipa{hEn}/ & \emph{hens} & /\textipa{hEnz}/ & \emph{hence} & /\textipa{hEns}/ \\
\emph{her} & /\textipa{h3\textrhoticity{}}/ & \emph{hers} & /\textipa{h3\textrhoticity{}z}/ & \emph{hearse} & /\textipa{h3\textrhoticity{}s}/ \\
\emph{how} & /\textipa{haU}/ & \emph{how's} & /\textipa{haUz}/ & \emph{house} & /\textipa{haUs}/ \\
\emph{in} & /\textipa{In}/ & \emph{inns}, \emph{ins} & /\textipa{Inz}/ & \emph{ince} & /\textipa{Ins}/ \\
\emph{jew} & /\textipa{dZu}/ & \emph{jew's}, \emph{jews} & /\textipa{dZuz}/ & \emph{juice} & /\textipa{dZus}/ \\
\emph{joy} & /\textipa{dZOI}/ & \emph{joys} & /\textipa{dZOIz}/ & \emph{joyce} & /\textipa{dZOIs}/ \\
\emph{julia} & /\textipa{dZulj2}/ & \emph{julia's} & /\textipa{dZulj2z}/ & \emph{julius} & /\textipa{dZulj2s}/ \\
\emph{knee} & /\textipa{ni}/ & \emph{knees} & /\textipa{niz}/ & \emph{niece} & /\textipa{nis}/ \\
\emph{knew} & /\textipa{nu}/ & \emph{news} & /\textipa{nuz}/ & \emph{noose} & /\textipa{nus}/ \\
\emph{law} & /\textipa{lO}/ & \emph{laws} & /\textipa{lOz}/ & \emph{los} & /\textipa{lOs}/ \\
\emph{lay} & /\textipa{leI}/ & \emph{lays} & /\textipa{leIz}/ & \emph{lace} & /\textipa{leIs}/ \\
\emph{may} & /\textipa{meI}/ & \emph{maize}, \emph{maze} & /\textipa{meIz}/ & \emph{mace} & /\textipa{meIs}/ \\
\emph{one} & /\textipa{w2n}/ & \emph{one's}, \emph{ones} & /\textipa{w2nz}/ & \emph{once} & /\textipa{w2ns}/ \\
\emph{pay} & /\textipa{peI}/ & \emph{pays} & /\textipa{peIz}/ & \emph{pace} & /\textipa{peIs}/ \\
\emph{peer} & /\textipa{pir}/ & \emph{peers} & /\textipa{pirz}/ & \emph{pierce} & /\textipa{pirs}/ \\
\emph{play} & /\textipa{pleI}/ & \emph{plays} & /\textipa{pleIz}/ & \emph{place} & /\textipa{pleIs}/ \\
\emph{ray} & /\textipa{reI}/ & \emph{raise}, \emph{rays} & /\textipa{reIz}/ & \emph{race} & /\textipa{reIs}/ \\
\emph{river} & /\textipa{rIv3\textrhoticity{}}/ & \emph{river's}, \emph{rivers} & /\textipa{rIv3\textrhoticity{}z}/ & \emph{reverse} & /\textipa{rIv3\textrhoticity{}s}/ \\
\emph{rye} & /\textipa{raI}/ & \emph{rise} & /\textipa{raIz}/ & \emph{rice} & /\textipa{raIs}/ \\
\emph{sin} & /\textipa{sIn}/ & \emph{sins} & /\textipa{sInz}/ & \emph{since} & /\textipa{sIns}/ \\
\emph{soar} & /\textipa{sOr}/ & \emph{sores} & /\textipa{sOrz}/ & \emph{source} & /\textipa{sOrs}/ \\
\emph{spy} & /\textipa{spaI}/ & \emph{spies} & /\textipa{spaIz}/ & \emph{spice} & /\textipa{spaIs}/ \\
\emph{true} & /\textipa{tru}/ & \emph{true's} & /\textipa{truz}/ & \emph{truce} & /\textipa{trus}/ \\
        \bottomrule
    \end{tabular}
    }
    \caption{Materials used in the forced-choice experiment. Token word embeddings are retrieved based on transcribed phonetic form (IPA given here). Corresponding orthographic forms from the LibriSpeech corpus consistent with these phonetic transcriptions are given for convenience.}
    \label{tbl:fc-materials}
\end{table*}

Below we address several possible confounds that arise from this experimental design.

\subsubsection{Possible confound: Bias for /z/}

The experimental items only contain \say{consistent} words ending in [z]. It is unfortunately impossible to design a fully balanced experiment under this structure due to the limitations of English phonotactics. Any hypothetical \say{consistent} word ending in [s] would have by construction a preceding voiceless sound (e.g. \emph{lips} [\textipa{lIps}]). However, the addition of a voiced consonant here ([\textipa{lIpz}]) is prohibited by more general constraints on English consonant cluster voicing.

At first glance, then, our results are also confounded with the simpler claim that these difference vectors function to add a [z] sound onto any base form. However, we have shown that the same vectors exhibit roughly similar performance in predicting inflected forms with [s] and [\textipa{Iz}] as with [z] in \Cref{sec:exp-allomorph} and \Cref{fig:transfer-allomorph} (compare results across columns).

\subsubsection{Possible confound: Vowel length}

English vowels are typically lengthened before voiced codas \citep{Zimmerman1958NoteOV,Peterson1960DurationOS}. This raises a second possible confound: the difference vectors might be capturing changes in vowel length due to voicing, rather than (or in addition to) the presence of the final sibilant. However, this explanation doesn’t fit with the results in \Cref{sec:exp-allomorph,fig:transfer-allomorph}. Consider the first column of that figure, which tests whether noun plurals ending in [s] can be predicted using difference vectors from various sources. The second cell (row 2, column 1) uses vectors from words ending in [z] to predict inflections of words ending in [s]. If vowel length were driving the effect, we’d expect worse performance here—since words ending in [z] would exhibit vowel lengthening but words ending in [s] would not. Yet performance is similar to the case where both source and target words end in [s] (row 1, column 1), suggesting vowel length mismatch isn’t a key factor here.

The regression results in \Cref{tbl:regression-results} support this same idea: the estimated effect of sound match (\texttt{allomorph_from=S × allomorph_to=S}) in the word probe model is only an average 0.08 improvement in rank score. 

\end{document}